%% file: hasvm.tex
\RequirePackage{fix-cm}
\documentclass[twocolumn]{svjour3}          % twocolumn
\input{shortcuts}

\input{acronyms}
\smartqed  % flush right qed marks, e.g. at end of proof
\usepackage{graphicx}
\usepackage{amsmath}
\usepackage{amssymb}
\usepackage{mathptmx}      % use Times fonts if available on your TeX system
\usepackage[round]{natbib}
\begin{document}

\title{Hierarchical Adaptive Structural SVM for Domain Adaptation%\thanks{Grants or other notes
%about the article that should go on the front page should be
%placed here. General acknowledgments should be placed at the end of the article.}
}
%\subtitle{Do you have a subtitle?\\ If so, write it here}

%\titlerunning{Short form of title}        % if too long for running head

\author{Jiaolong Xu         \and
        Sebastian Ramos \and
        David V{\'a}zquez \and
        Antonio M. L{\'o}pez %etc.
}

%\authorrunning{Short form of author list} % if too long for running head

\institute{
			J. Xu, S. Ramos, A. M. L{\'o}pez \at
			Computer Vision Center, Barcelona, Spain\\
			Dept. of Computer Science, Universitat Aut{\'o}noma de Barcelona, \\
            Barcelona, Spain\\
            \email \{{jiaolong, sramosp, antonio\}@cvc.uab.es}           %  \\
			\and
            D. V{\'a}zquez \at
            Computer Vision Center, Barcelona, Spain\\
            \email{dvazquez@cvc.uab.es}           %  \\
%             \emph{Present address:} of F. Author  %  if needed
}

\date{Received: date / Accepted: date}
% The correct dates will be entered by the editor

\maketitle

\begin{abstract}
A key topic in \emph{classification} is the accuracy loss produced when the data distribution in the training (\emph{source}) domain differs from that in the testing (\emph{target}) domain. This is being recognized as a very relevant problem for many computer vision tasks such as image classification, object detection, and object category recognition. In this paper, we present a novel \emph{domain adaptation} method that leverages multiple target domains (or sub-domains) in a hierarchical adaptation tree. The core idea is to exploit the commonalities and differences of the jointly considered target domains.

Given the relevance of structural SVM (SSVM) classifiers, we apply our idea to the adaptive SSVM (A-SSVM), which only requires the target domain samples together with the existing source-domain classifier for performing the desired adaptation. Altogether, we term our proposal as hierarchical A-SSVM (HA-SSVM).   

As proof of concept we use HA-SSVM for pedestrian detection and object category recognition. In the former we apply HA-SSVM to the deformable part-based model (DPM) while in the latter HA-SSVM is applied to multi-category classifiers. In both cases, we show how HA-SSVM is effective in increasing the detection/recognition accuracy with respect to adaptation strategies that ignore the structure of the target data. Since, the sub-domains of the target data are not always known a priori, we shown how HA-SSVM can incorporate sub-domain structure discovery for object category recognition.
\end{abstract}

\section{Introduction}

Besides the data representations and learning algorithms used in classification tasks, other relevant fact that has been increasingly considered within this context is the so denominated \emph{dataset bias}. This is a very common problem in real-world classification applications that makes the classifiers suffer from loss in accuracy when the data distribution in the training (\emph{source}) domain differs from that in the testing (\emph{target}) domain. In order to face this \emph{domain adaptation} challenge a variety of methods have been increasingly explored within the machine learning field \citep{Daume:2007, Yang:2007,  Mansour:2008, Ben:2009, Duan:2009b, Pan:2009}  and most recently within the computer vision one \citep{Bergamo:2010, Saenko:2010, Gopalan:2011, Duan:2012, Vazquez:2012, Vazquez:2014, Hoffman:2012, Hoffman:2013, kan_ijcv:2014} since image classification, visual object detection and object category recognition are tasks where dataset bias is usually relevant.

\begin{figure*}[!t]
\centering
\centerline{\includegraphics[width=0.95\textwidth]{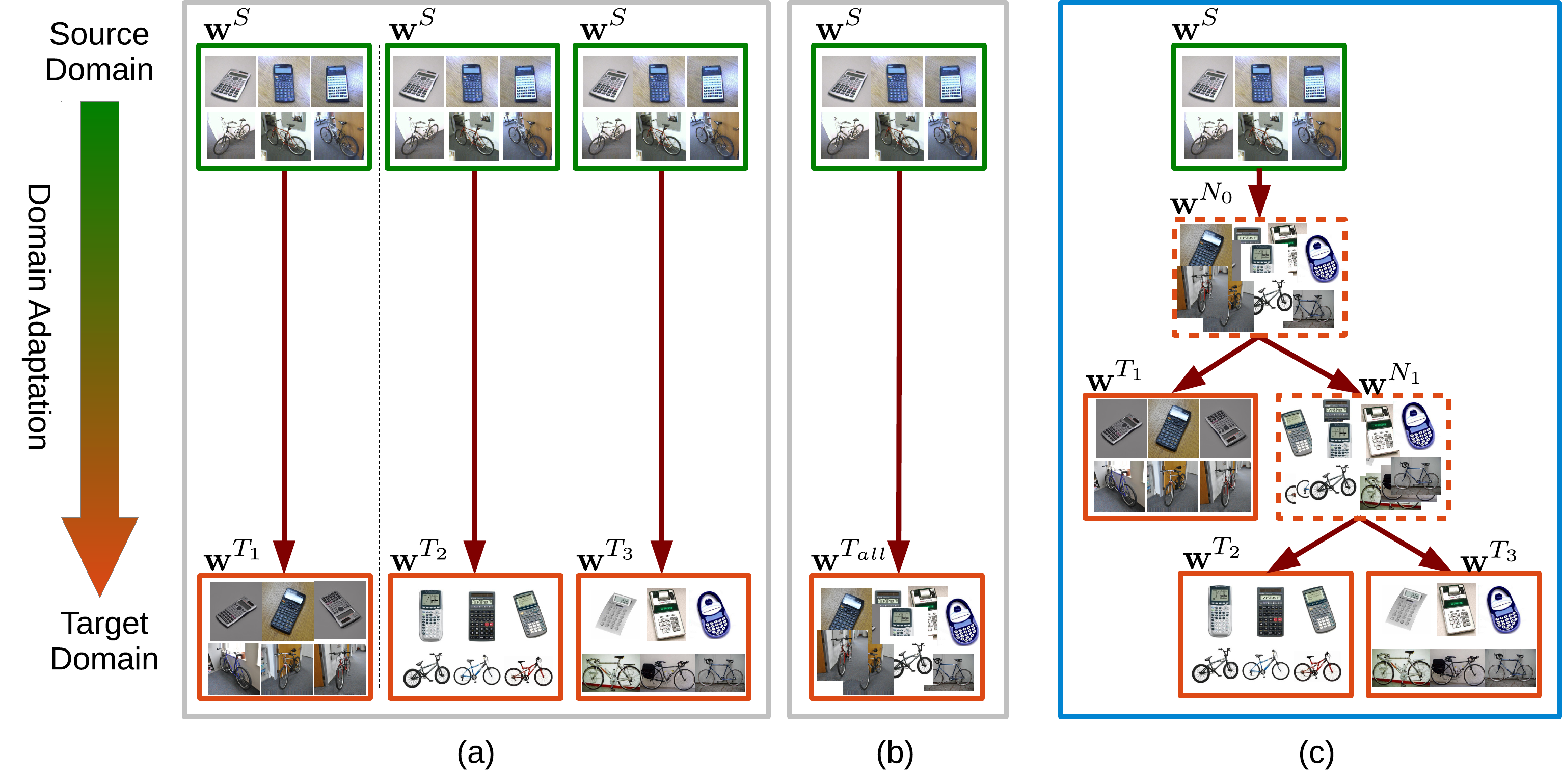}}
\caption{Domain adaptation methods: without losing generality we assume a single source domain and three correlated target domains (three different datasets depicting the same object categories). (a) Single layer domain adaptation: adapting to each target domain $\textbf{w}^{T_j}$ independently. (b) Single layer domain adaptation: pooling multiple target domains. (c) Proposed hierarchical multi-layer domain adaptation. The target domains are organized in an adaptation tree. Adaptation to intermediate nodes allows to exploit commonalities between children sub-domains, while adaptation to final sub-domains allows to consider their differences. Each path from the root to a leaf of the hierarchy can be thought as a progressive adaptation, but all models (intermediate and final) are learned jointly.}
\label{fig:Model}
\end{figure*}

Many domain adaptation methods assume a single domain shift between the data, {\ie}, they perform the adaptation from a single source domain to a single target domain \citep{Daume:2007, Bergamo:2010, Saenko:2010, Duan:2012, Vazquez:2012, Vazquez:2014, Hoffman:2013, kan_ijcv:2014}. Some others consider multiple source domains \citep{Yang:2007, Mansour:2008, Duan:2009b, Gopalan:2011, Hoffman:2012} and propose to leverage labeled data from them to perform the domain adaptation, {\ie}, the underlying idea is to cover as much variability as possible at the source level for making more accurate predictions given a partially new domain (the target). In this paper we focus on the complementary case to these works. In other words, the main novelty is the study of the effectiveness of domain adaptation when we can structure the target domain as a hierarchy ({\eg}, leveraging multiple correlated target domains or using some criteria to build sub-domain partitions). Moreover, due to its practical implications, we focus on methods that do not require to revisit the source data for the adaptation.   

The main idea of our approach is illustrated in \figurename~\ref{fig:Model}. Without losing generality assume that we have a prior source model $\textbf{w}^S$ ({\eg} a SVM hyperplane) and we would like to adapt it to multiple target domains ($T_1,T_2,T_3$) from which we have labeled data. Traditionally, $\textbf{w}^S$ is adapted to each target domain separately, as illustrated in (a). Other option is to pool multiple target domains into a single one and adapt $\textbf{w}^S$ to a mixed target domain as in (b). We refer to these strategies as single-layer domain adaptation. 

Instead of performing isolated single-layer adaptations, we propose to make use of the relatedness of the target domains while exploiting their differences. Concretely, as it is presented in (c), we organize multiple target domains into a hierarchical structure (tree) and adapt the source model to them jointly. The adaptation to intermediate nodes allows to exploit commonalities between children sub-domains ({\eg}, approach (b) is considered thanks to the root node of the hierarchy), while the adaptation to the final sub-domains allows to consider their differences. 

Each path from the root to a leaf of the hierarchy can be thought as a progressive adaptation. However, as we will see, the adaptation of the whole hierarchy is done at once under the same objective function. This implies that our adaptation strategy is also useful in cases where the labeled data from the target domain is scarce but at the same time presents certain variability (sub-domains) worth to consider. Note that by using the approach in (a) such a reduced target domain dataset would be divided into even smaller target sub-domain datasets, which in general would end up in a poorer adaptation. On the other hand, following (b) the potential target sub-domains would be just ignored. 

Given the widely use of SVM classifiers, we focus on the model-transform-based domain adaptation method known as {\em adaptive SVM} (A-SVM) \citep{Yang:2007}. A-SVM does not require source domain samples, only target domain ones, which can significantly reduce the training (adaptation) time. In fact, since we will address problems requiring {\em structural SVM} (SSVM), we will use the A-SVM variation for SSVM that we introduced in \citep{Xu_PAMI:2014}, namely the {\em adaptive SSVM} (A-SSVM). Altogether we term our approach as {\em hierarchical A-SSVM} (HA-SSVM).   

%In some cases, the latent domains of the target data may not be explicitly given, {\eg}, a collection of web searched images can come from many different domains or an on-board/hand-held camera may capture images from different continuously changing scenarios. Therefore, discovering such sub-domains is required for applying our domain adaptation algorithm. Moreover, the situation can correspond to a totally unsupervised scenario where the class/category labels of the target data are neither available. Accordingly, we show how HA-SSVM can incorporate a domain reshaping technique \citep{Gong:2013} for discovering the latent sub-domain structure in such an unsupervised case.

As proof of concept we apply our method in two problems of paramount importance within the computer vision field, namely pedestrian detection and object category recognition, the former implies to use HA-SSVM with the wide\-spread deformable part-based model (DPM) while the latter implies to use HA-SSVM with multi-category classifiers. In both cases, we will show how HA-SSVM is effective in increasing the detection/recognition accuracy with respect to state-of-the-art strategies that ignore the structure of the target data. Moreover, focusing on the object category recognition application, we will also evaluate HA-SSVM in an scenario were the target sub-domains are not available a priori and must be discovered.  

The rest of the paper is organized as follows. Section \ref{sec:RelatedWork} overviews the related work. In section \ref{sec:Method} we detail the general formulation of the proposed approach and its optimization method, as well as how to incorporate domain reshaping for discovering latent domains. Sections \ref{sec:DA_PedestrainDetetcion} and \ref{sec:DA_ObjectRecognition} present the experimental results of HA-SSVM for pedestrian detection and object recognition, respectively. Finally, section \ref{sec:Conclusion} draws the conclusions and future research lines.

\section{Related Work}
\label{sec:RelatedWork}
Despite the variety of domain adaptation methods proposed in the last decades--see \citep{Jiang:2008} for a comprehensive overview--, in computer vision, current methods can be broad\-ly categorized in two main groups, namely feature-transform-based methods and model-transform-based methods.

Feature-transformation-based methods attempt to learn a transformation matrix/kernel over the feature space of different domains, and then apply a classifier \citep{Saenko:2010, Kulis:2011, Gopalan:2011, hoffman_IJCV:2014, gong_ijcv:2014}. For instance, the max-margin domain transforms method \citep{Hoffman:2013} jointly learns a feature transformation and a discriminative classifier via multi-task learning. On the other hand, model-transform-based approaches concentrate on adapting the parameters of the classifiers, often SVM, including: weighted combination of source and target SVMs, transductive SVM \citep{Bergamo:2010, Vazquez:2012}, feature replication \citep{Daume:2007, Vazquez:2014}, and regulari\-zation-based methods as A-SVM \citep{Yang:2007}, its successor the projective model transfer SVM (PMT-SVM) \citep{Aytar:2011} and its variant A-SSVM \citep{Xu_PAMI:2014}.  

Among these methods, the SVM regularization-based ones have a significant advantage as they do not require revisiting source domain data for the adaptation. This would be favorable for many domain adaptation tasks in computer vision, since the source datasets are typically large and computing the features is expensive. Besides, it can even handle the case where the source data is missing at the moment of the adaptation. Basically, these methods learn the target classifier $f^T(\textbf{x})$ by adding a perturbation function $\Delta f(\textbf{x})$ to the source classifier $f^S(\textbf{x})$ so that $f^T(\textbf{x}) = f^S(\textbf{x}) + \Delta f(\textbf{x})$. Our approach can be regarded as a higher level model of A-SSVM, as it considers the structural relation between different domains and integrates multiple A-SSVM adaptations in a hierarchical model.

In the context of domain adaptation between multiple domains, several methods close to our work have been proposed in the natural language processing (NLP) community \citep{Finkel:2009,Daume:2009}, which are Bayesian-based approaches. While for multi-domain adaptation most of the focus is on multiple sources, little attention is paid on the relation of multiple target domains. Our domain adaptation method aims to leverage multiple target domains by considering their hierarchical structural relation. 

Most of the domain adaptation algorithms are validated assuming that the underlying domains are well-defined. However, multiple unknown domains may exist \citep{Hoffman:2012}. In fact, in some cases image data is difficult to manually divide into discrete domains required by adaptation algorithms \citep{Gong:2013}. In \citep{Hoffman:2012}, a sub-domain discovery algorithms is proposed, it focuses on discovering multiple hidden source domains. The most recent work of \citep{Gong:2013} can discover domains among both training and testing data, which benefits existing multi-domain adaptation algorithms. In this paper, we also include experiments where HA-SSVM is applied to discovered target sub-domains for object category recognition.

The vision tasks where our method can be applied are several, however, in this paper our experiments focus on cross-domain multi-category object recognition and pedestrian detection based on the deformable part-based model (DPM). The former has been a benchmark for the proof of most domain adaptation methods developed for vision tasks \citep{Saenko:2010,Kulis:2011,Gong:2012}. Despite its relevance, the latter has been just rarely addressed in the literature \citep{Vazquez:2012, Vazquez:2014, Xu_PAMI:2014} from the viewpoint of domain adaptation. However, it introduces the interesting challenges of dealing with rather unbalanced classes ({\ie}, pedestrians {\vs} background). 

\section{Proposed Method}
\label{sec:Method}
\subsection{General Model}
Our proposal is illustrated in \figurename~\ref{fig:Model}. Assume we have a prior model $\textbf{w}^S$ from the source domain $\mathcal{D}^{S}$ and multiple target domains $\mathcal{D}^{T_j}, j \in [1, D]$. Traditionally, $\textbf{w}^S$ is adapted to each target domain independently, as illustrated in (a), or to the pooled target domain as in (b), which we call single-layer domain adaptation in this paper. In contrast, we propose to make use of the relatedness of multiple target domains by combining them into a hierarchical adaptation tree, and adapt the prior model to them hierarchically, as in (c).

The proposed hierarchical model can be applied to any supervised learning algorithm which can incorporate prior information. In this work, we focus on the widely used SVM. This learning method considers a loss term $\mathcal{L}(\textbf{w}; \mathcal{D})$ that captures the error with respect to the training data $\mathcal{D}$ and a regularization term $\mathcal{R}(\textbf{w})$ that penalizes model complexity. In fact, we will focus on domain adaptation with structural SVM (SSVM), giving rise to our hierarchical A-SSVM (HA-SSVM) in Sect \ref{ssec:HA-SSVM}. 
%Additionally, we further consider the scenario where no prior structure knowledge of the target domains is given, and we incorporate the domain reshaping method \citep{Gong:2013} to discover and build hierarchical latent domains in Sect \ref{ssec:Method_Domain_Clustering}.

%---------------------------------------------------------------------------------------
\subsection{Domain Adaptation Methods}
\label{ssec:HA-SSVM}
For the sake of a better understanding, in this subsection, we introduce the involved concepts by progressive order of complexity. In Sect. \ref{sssec:A-SVM} we focus on single-layer domain adaptation based on adaptive SVM (A-SVM). Then, in Sect. \ref{sssec:HA-SVM} we develop our hierarchical A-SVM (HA-SVM) model. We show how to learn its parameters by using a multiple task learning (MTL) paradigm. Finally, in Sect. \ref{sssec:HA-SSVM} we consider SSVM and, therefore, introduce HA-SSVM. 

\subsubsection{Adaptive SVM (A-SVM)}
\label{sssec:A-SVM}
A-SVM is a model-transform-based method, which adapts the model parameters from the source $\mathcal{D}_{l}^{S}$ to the target domain $\mathcal{D}_{l}^{T}$ ($l$ indicates that the samples are labeled). Given the source domain model $\textbf{w}^S$, the target domain model $\textbf{w}^T$  is learned by minimizing the following objective function:
\begin{equation}
\label{eq:ASVM}
\min_{\textbf{w}^T} \dfrac{1}{2} \|\textbf{w}^T - \textbf{w}^{S}\|^{2} + C \mathcal{L}(\textbf{w}^T; \mathcal{D}_{l}^{T}),
\end{equation}
where the regularization term $\|\Delta\textbf{w}\|^{2} = \|\textbf{w}^T - \textbf{w}^{S}\|^{2}$ constrains the target model $\textbf{w}^T$ to be close to the source one $\textbf{w}^S$. Eq.~(\ref{eq:ASVM}) is also called one-to-one domain adaptation. At the testing time, we apply the following decision function to the target domain:
\begin{equation}
\label{eq:ASVM_f}
f^T(\textbf{x}) = {\textbf{w}^T}'\Phi(\textbf{x}) = f^S(\textbf{x}) + \Delta{\textbf{w}}'\Phi(\textbf{x}),
\end{equation}
where $\Phi(\textbf{x})$ is the feature vector for target domain sample $\textbf{x}$ and $f^S(\textbf{x})$ is the output score from the source domain classifier. Thus, A-SVM is essentially learning a perturbation function $\Delta{f(\textbf{x})} = \Delta{\textbf{w}}'\Phi(\textbf{x})$ based on the source classifier.

\subsubsection{Hierarchical Adaptive SVM (HA-SVM)}
\label{sssec:HA-SVM}
For the sake of a more understandable explanation but without losing generality, we give the formulation of HA-SVM for a hierarchy of three layers as the one illustrated in \figurename~\ref{fig:Model} (c). Assume we have a source model $\textbf{w}^{S}$ and three target domains $\textbf{w}^{^{T_j}}$, $j \in [1, 3]$. Let $\textbf{w} = [{\textbf{w}^{N_0}}', {\textbf{w}^{N_1}}', {\textbf{w}^{T_1}}', {\textbf{w}^{T_2}}', {\textbf{w}^{T_3}}']'$, then the objective function of the three-layers HA-SVM is written as follows:
\begin{equation}
\label{eq:H-ASVM}
\begin{array}{ll}
J(\textbf{w}) = &\dfrac{1}{2} \|\textbf{w}^{N_0} - \textbf{w}^{S}\|^{2} + C \sum_{j=1}^{3} \mathcal{L}(\textbf{w}^{N_0}; \mathcal{D}_{l}^{T_j})\\
\vspace{0.2cm}
&+\dfrac{1}{2} \|\textbf{w}^{N_1} - \textbf{w}^{N_0}\|^{2} + C \sum_{j=2}^{3} \mathcal{L}(\textbf{w}^{N_1};\mathcal{D}_{l}^{T_j})\\
\vspace{0.2cm}
&+\dfrac{1}{2} \|\textbf{w}^{T_1} - \textbf{w}^{N_0}\|^{2} + C \mathcal{L}(\textbf{w}^{T_1};\mathcal{D}_{l}^{T_1})\\
\vspace{0.2cm}
&+\dfrac{1}{2} \|\textbf{w}^{T_2} - \textbf{w}^{N_1}\|^{2} + C \mathcal{L}(\textbf{w}^{T_2};\mathcal{D}_{l}^{T_2})\\
\vspace{0.2cm}
&+\dfrac{1}{2} \|\textbf{w}^{T_3} - \textbf{w}^{N_1}\|^{2} + C \mathcal{L}(\textbf{w}^{T_3};\mathcal{D}_{l}^{T_3})\\
\end{array}
\end{equation}
\Eq{eq:H-ASVM} is in a multi-task learning paradigm form, where the optimization of each $\textbf{w}^{T_j}$ can be understood as an individual task. The key issue of the multi-task learning lies in how the relationships between tasks are incorporated. As we can see from \Eq{eq:H-ASVM}, each task is related by the regularization term, {\eg}, $T_2$ and $T_3$ are connected by $\|\textbf{w}^{T_j}-\textbf{w}^{N_1}\|^{2}$, while $T_1$ is directly connected to $N_0$, which is adapted from $\textbf{w}^{S}$. 

At testing time, for a testing sample from target domain $j$, we can directly extract the learned parameters $\textbf{w}^T_j$ and apply the linear decision function:
\begin{equation}
\label{eq:HASVM_f}
f^{T_j}(\textbf{x}) = {\textbf{w}^{T_j}}' \Phi(\textbf{x})~.
\end{equation}

Comparing to the single-layer adaptation $\|\textbf{w}^{T_j}-\textbf{w}^{S}\|^{2}$ as in \figurename~\ref{fig:Model} (a), HA-SVM has several advantages. First, HA-SVM can make use of training samples from multiple related target domains instead of just one. For example, a single-layer domain adaptation only uses the training samples from $T_j, j\in \{1,2, 3\}$ in three different optimization runs, while HA-SVM can integrate the samples from the three target domains accounting for their hierarchical structure. 
%Assume $T_1$ does not contain sufficient samples for adaptation, it can {\em borrow} examples from $T_2$ and $T_3$. 
Second, the target model $\textbf{w}^{T_j}$ is not directly regularized by $\textbf{w}^{S}$ but some shared intermediate models $\textbf{w}^{N_i}$, which allows $\textbf{w}^{T_j}$ to be regularized in a more flexible space. As $\textbf{w}^{T_j}$ goes down apart from $\textbf{w}^{S}$ further in the adaptation tree, less constrain from $\textbf{w}^{S}$ is imposed. This can be interpreted as a progressive adaptation. 

For single-layer domain adaptation, another straightforward strategy is to pool all target domains and train a single adaptive SVM with all available target samples, as illustrated in \figurename~\ref{fig:Model} (b). Comparing to this method, HA-SVM can take the same advantage of using all available labeled data while allows each target domain model to be more discriminative in its own domain. The pooling-based method requires the final model to compromise to each domain in order to minimize the training error, and thus such model may lose the discriminative power in the individual target domains. Our experimental results in {\sect{sec:Experiments_DPM} and \sect{sec:Experiments_ObjRec}} confirm this observation.

To minimize Eq.~(\ref{eq:H-ASVM}), we employ Quasi-Newton LBFGS method, %\citep{LBFGS}.
which requires the objective function and the partial derivatives of its parameters. These partial derivatives are:
\begin{equation}
\label{eq:LBFGS1}
\begin{array}{ll}
\frac{\partial J(\textbf{w})}{\partial \textbf{w}^{N_0}} =& 3\textbf{w}^{N_0} - \textbf{w}^{S} - \textbf{w}^{N_1} - \textbf{w}^{T_1} + C \sum_{j=1}^{3} \frac{\partial \mathcal{L}(\textbf{w}^{N_0}; \mathcal{D}_{l}^{T_j})}{\partial \textbf{w}^{N_0}}~,\\

\frac{\partial J(\textbf{w})}{\partial \textbf{w}^{N_1}} =& 3\textbf{w}^{N_1} - \textbf{w}^{N_0} - \textbf{w}^{T_1} - \textbf{w}^{T_2} + C \sum_{j=2}^{3} \frac{\partial \mathcal{L}(\textbf{w}^{N_1}; \mathcal{D}_{l}^{T_j})}{\partial \textbf{w}^{N_1}}~,\\

\frac{\partial J(\textbf{w})}{\partial \textbf{w}^{T_1}} =&  \textbf{w}^{T_1}-\textbf{w}^{N_0} + C \frac{\partial \mathcal{L}(\textbf{w}^{T_1}; \mathcal{D}_{l}^{T_1})}{\partial \textbf{w}^{T_1}}~,\\

\frac{\partial J(\textbf{w})}{\partial \textbf{w}^{T_2}} =& \textbf{w}^{T_2}-\textbf{w}^{N_1} + C \frac{\partial \mathcal{L}(\textbf{w}^{T_2}; \mathcal{D}_{l}^{T_2})}{\partial \textbf{w}^{T_2}}~,\\

\frac{\partial J(\textbf{w})}{\partial \textbf{w}^{T_3}} =& \textbf{w}^{T_3}-\textbf{w}^{N_1} + C \frac{\partial \mathcal{L}(\textbf{w}^{T_3}; \mathcal{D}_{l}^{T_3})}{\partial \textbf{w}^{T_3}}~.
\end{array}
\end{equation}

In our implementation, the LBFGS based optimization converges to the optimum efficiently for both single-layer A-SVM and HA-SVM (as well as for the HA-SSVM defined in next subsection).

\subsubsection{Hierarchical Adaptive Structured SVM (HA-SSVM)}
\label{sssec:HA-SSVM}
The proposed HA-SVM can be extended for SSVM, giving rise to our HA-SSVM. SSVM allows the training of a classifier for general structured output labels. SSVM minimizes the following regularized risk function:
\begin{equation}
 \label{ssvm_obj}
 \begin{array}{ll}
 \min_{\textbf{w}} \dfrac{1}{2} \|\textbf{w}\|^{2} \\
 + C\sum_{i=1}^N [\max_{y} \textbf{w}'\Phi{(\textbf{x}_{i}, y)} + \Delta(y_{i},y) - \Phi{(\textbf{x}_{i}, y_{i})}]~,
 \end{array}
\end{equation}
where $y_{i}$ is the ground truth output (label) of sample $\textbf{x}_{i}$, and $y$ runs on the alternative outputs. $\Delta(y_{i},y)$ is a distance in output space. $\Phi(\textbf{x}, y)$ is the feature vector from a given sample $\textbf{x}$ of label $y$. Accordingly, Eq.~(\ref{eq:ASVM}) can be extended to A-SSVM as follows: 
% Explicit form
\begin{equation}
 \label{eq:ASSVM}
 \begin{array}{ll}
 \min_{\textbf{w}^T} \dfrac{1}{2} \|\textbf{w}^T - \textbf{w}^S\|^{2} \\
 + C\sum_{i=1}^N [\max_{y} {\textbf{w}^T}'\Phi{(\textbf{x}_{i}, y)} + \Delta(y_{i},y) - \Phi{(\textbf{x}_{i}, y_{i})}]~.
 \end{array}
\end{equation}
Correspondingly, the final adapted classifier $f^{T}$ can be written as:
\begin{equation}
\label{assvm_define}
f^{T}(\textbf{x}) = \max_{y} [{\textbf{w}^{S}}'\Phi{(\textbf{x}, y)} + \underbrace{\Delta{\textbf{w}}'\Phi{(\textbf{x}, y)}}_{\Delta f(\textbf{x})}]~,
\end{equation}
where $\Delta{\textbf{w}} = \textbf{w}^T - \textbf{w}^S$. Therefore, Eq.~(\ref{eq:ASSVM}) can be integrated into the proposed hierarchical adaptation framework descri\-bed in \Sect{sssec:HA-SVM}. In particular, we must proceed in the same way than going from (\ref{eq:ASVM}) to (\ref{eq:H-ASVM}), but now starting with (\ref{eq:ASSVM}), thus, giving rise to HA-SSVM.

\section{Domain Adaptation of DPMs}
\label{sec:DA_PedestrainDetetcion}
In this section we apply HA-SSVM in the popular deformable part-based model (DPM) framework \citep{Felzenszwalb:2010}, focusing on pedestrian detection. The DPM learns a linear classification parameter $\textbf{w}$, which parametrizes a set of parts and deformations, to decide whether a detection window contains a pedestrian or background. The learning of a DPM is usually formulated as the latent SVM (LSVM) framework \citep{Felzenszwalb:2010}. However, it is also possible to use a latent structural SVM (LSSVM) formulation \citep{Zhu:2010,Girshick:2012}, which can be solved by the Convex-Concave Procedure (CCCP). The LSSVM form of a DPM objective function can be written as:
\begin{equation}
\label{eq:DPM}
\begin{array}{ll}
\min_{\textbf{w}} \dfrac{1}{2} \|\textbf{w}\|^{2}&+ C\sum_{i=1}^{N} [ \max_{\textbf{h}}(\textbf{w}'\Phi(\textbf{x}_i,\textbf{h}) \\
&+ \Delta(y, y_i)) - \textbf{w}'\Phi(\textbf{x}_i, \textbf{h}^*)],
\end{array}
\end{equation}
where $\textbf{h}$ is the latent variable which defines the object hypothesis, {\eg}, the alignment of parts, $\Phi(\textbf{x},\textbf{h})$ concatenates HOG-based \citep{Dalal:2005,Felzenszwalb:2010} appearance features and part spatial alignment features, $\Delta(y, y_i)$ is the $0$-$1$ loss function that returns $0$ if the binary label $y$ equals $y_i$, and $1$ otherwise. $\textbf{h}^*$ plays the role of ground truth output of example $i$ and it is computed at the concave stage of CCCP. We refer to \citep{Zhu:2010} \citep{Girshick:2012} for more details. A-SSVM can be applied to DPM according to \Eq{eq:ASSVM}. For example, A-SSVM for DPM can be written as:
\begin{equation}
\label{eq:asvm_dpm}
\begin{array}{ll}
\min_{\textbf{w}} \dfrac{1}{2} \|\textbf{w}^T - \textbf{w}^S\|^{2}&+ C\sum_{i=1}^{N} [ \max_{\textbf{h}}({\textbf{w}^T}'\Phi(\textbf{x}_i,\textbf{h}) \\
&+ \Delta(y, y_i)) - {\textbf{w}^T}'\Phi(\textbf{x}_i, \textbf{h}^*)].
\end{array}
\end{equation}
Thus, by applying \Eq{eq:H-ASVM}, we can build HA-SSVM for DPM. We implemented it in the DPM 5.0 framework \citep{DPM:release5}, which is the latest at the moment of doing this research. When applying an adapted DPM in a particular target domain ({\ie}, in testing time), we do not use the full vector of parameters jointly learned for all the hierarchy of target domains, instead we only use the sub-vector of parameters corresponding to such particular target domain. In other words, we follow \Eq{eq:HASVM_f}.
%-----------------------------------------------------------------------
\begin{figure}
(a)\includegraphics[width=0.5\columnwidth]{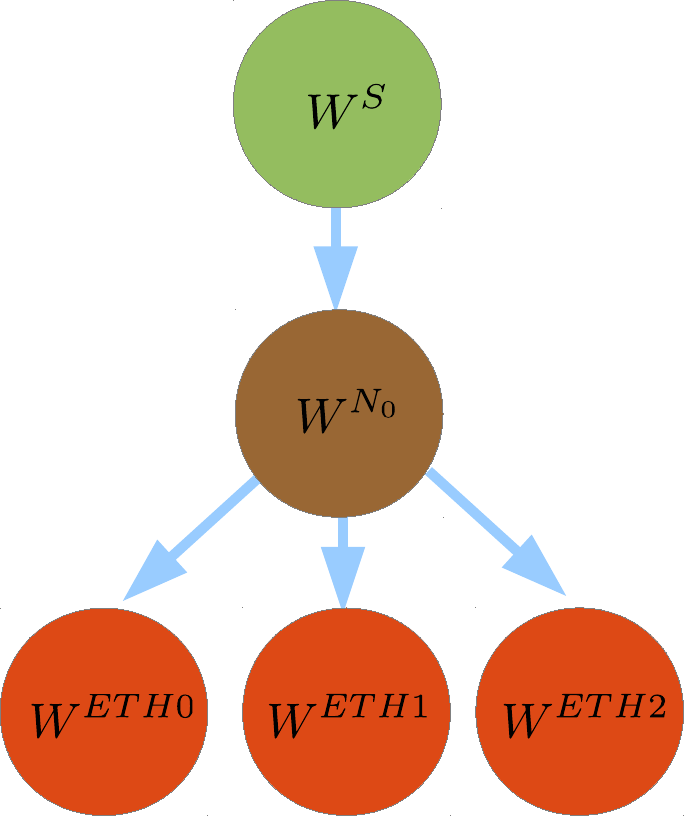}\hspace{0.0\columnwidth}
(b)\includegraphics[width=0.41\columnwidth]{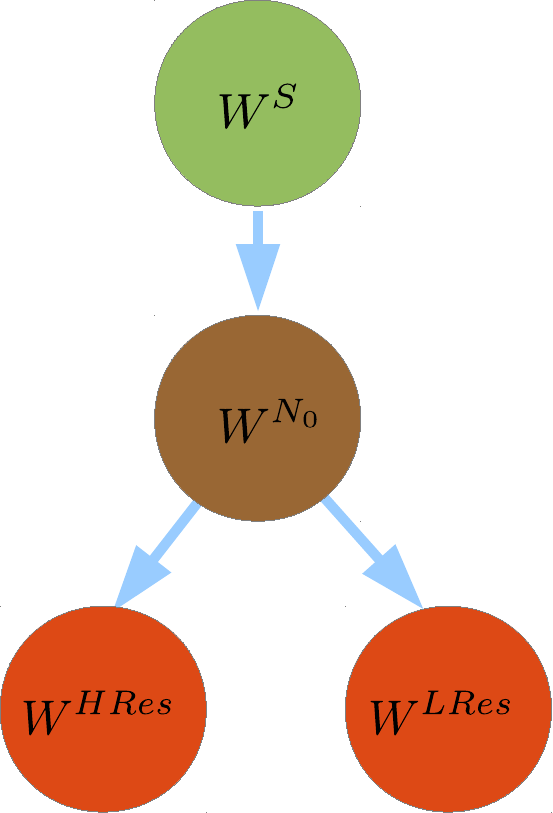}\\
\caption{HA-SSVM applied to DPM: (a) adaptation to three related datasets (domains), namely ETH0, ETH1 and ETH2, which were acquired with the same camera but different environments; (b) adaptation of a single resolution detector for applying different detectors when processing small and large image windows, which would correspond to pedestrians imaged with low (LRes) and high (HRes) resolution respectively.}
\label{fig:AdaptationTree}
\end{figure}
%----------------------------------------------------------------------------------------------------------------------------

%----------------------------------------------
\begin{center}
\begin{figure}
\includegraphics[width=\columnwidth, trim = 0 0 250 0, clip]{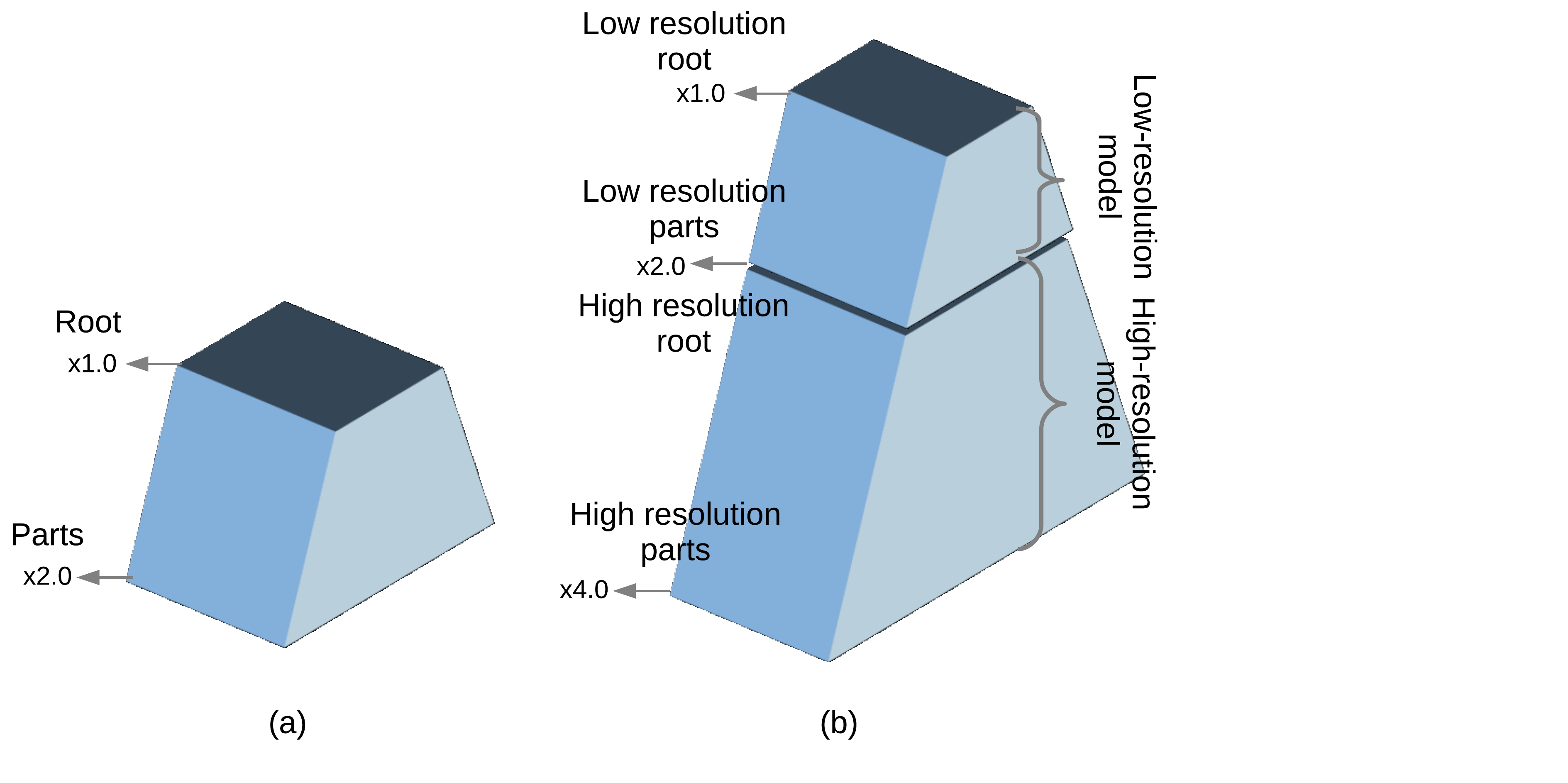}
\caption{(a) Original feature pyramid in DPM. (b) The extended feature pyramid for multi-resolution adaptive DPM.}
\label{fig:hasvm_multires}
\end{figure}
\end{center}

%----------------------------------------------
%\textcolor{red}{\textbf{TODO: Add a figure for illustration.}}.
%=============================================================================================================================
\subsection{Experiments on Pedestrian Detection}
\label{sec:Experiments_DPM}
\figurename~\ref{fig:AdaptationTree} illustrates two different cases of DPM domain adaptation using HA-SSVM that we evaluate here. In (a) the source classifier is adapted to three different target domains (different datasets in this case). In (b) we adapt the source classifier to detect pedestrians from image windows of two different resolution categories.  The main idea is to divide the target domain into sub-domains according to the resolution of the pedestrian samples, {\ie}, different resolutions are regarded as different domains. Here we consider only two resolutions, low and high. Note that low resolution pedestrians tend to be blur and their poses are less discriminative than for high resolution ones.  

\begin{figure}[t]
\includegraphics[width=8.5cm]{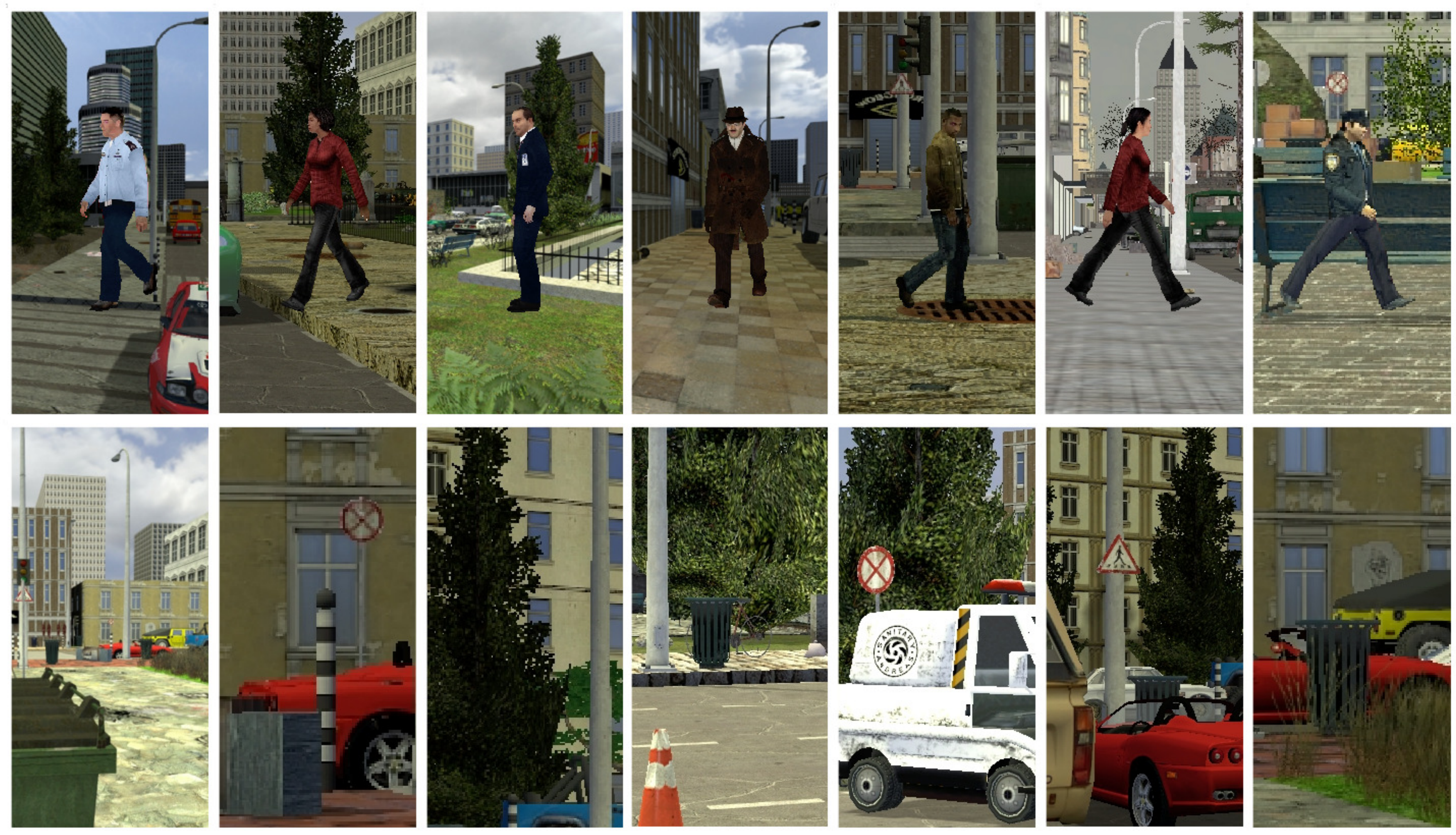}
\caption{Virtual-world pedestrians and background images.}
\label{fig:virtual_examples}
\end{figure}

%----------------------------------------------------------------------------------
\subsubsection{Datasets} As source-domain virtual-world dataset\footnote{It is publicly available under the name {\em CVC-07 DPM Virtual-World Pedestrian Dataset} at {\tt http://www.cvc.uab.es/adas}} (\fig{fig:virtual_examples}) we use the same as in \citep{Vazquez:2014, Xu_PAMI:2014, Xu_ITS:2014}. The target-domain real-world datasets that we use are ETH \citep{Ess:2007}, Caltech \citep{Dollar:2012} and KITTI \citep{Geiger:2012}. The ETH dataset consists of three sub-datasets, namely ETH0, ETH1 and ETH2, and is used to evaluate the setting of \figurename~\ref{fig:AdaptationTree}(a). These sub-datasets are collected from different environments but with the same camera sensor and pose, thus, we consider them as related sub-domains. Caltech and KITTI are used in two different experiments, both for evaluating the setting of \figurename~\ref{fig:AdaptationTree}(b). 
%----------------------------------------------------------------------------------

\begin{center}
\begin{table}[t]
{\scriptsize \input{table_learned_classifiers_dpm.tex}}
\caption{Different types of learned DPM classifiers.}
\label{tb_learned_classifiers_dpm}
\end{table}
\end{center}

%-------------------------------------------------------------------------------
\begin{figure}
\begin{minipage}[b]{1.0\linewidth}
  \centering
  \centerline{\includegraphics[width=8.5cm]{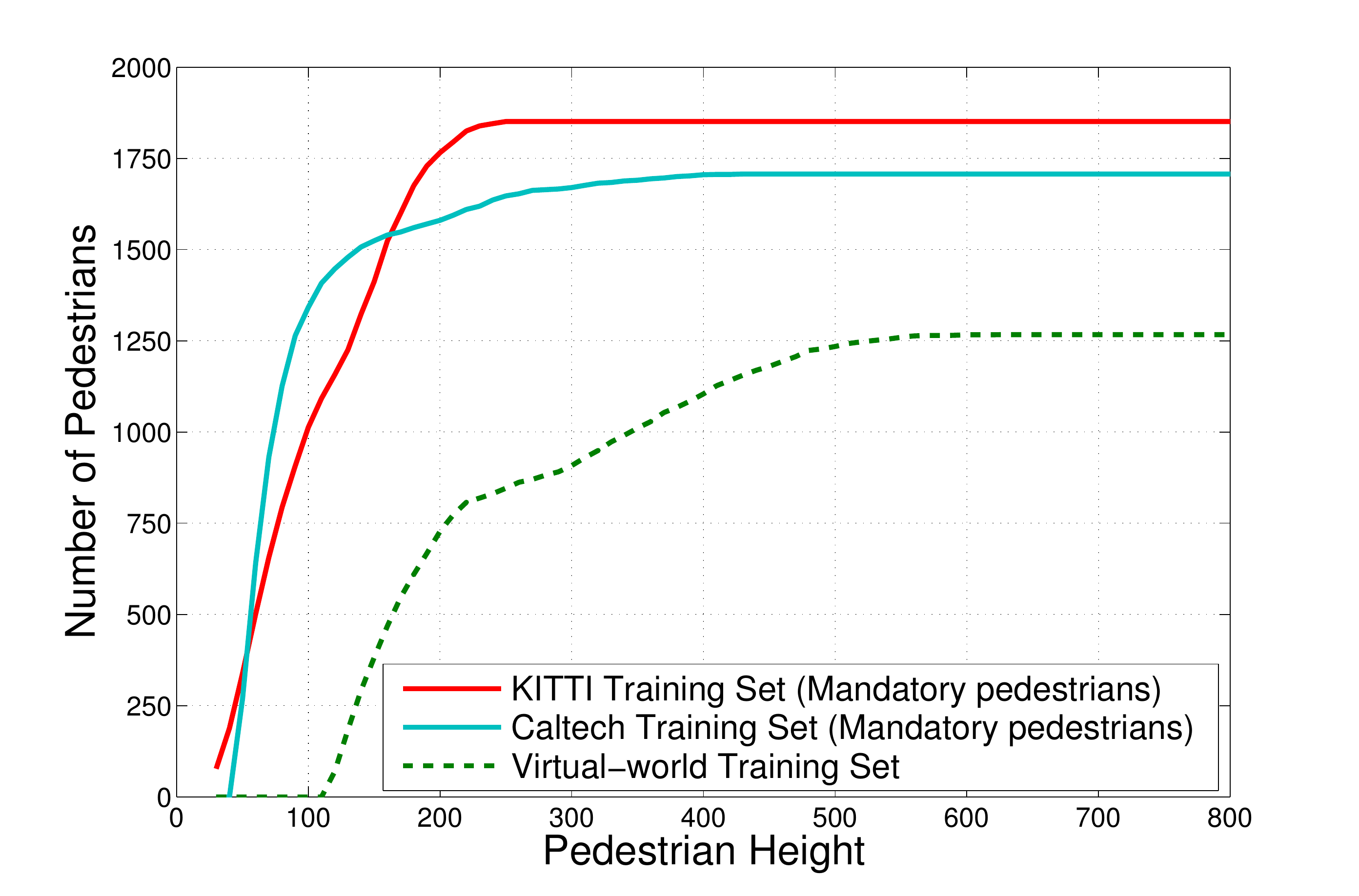}}
  %\vspace{1.5cm}
  % \centerline{Adaptation Tree}\medskip
\end{minipage}
\caption{Cumulative histogram of the pedestrians' height in Caltech,  KITTI and virtual-world training dataset. The virtual-world dataset contains less low-resolution pedestrians than the real-world ones.}
\label{fig:CHis}
\end{figure}
%-------------------------------------------------------------------------------

\begin{figure*}
\begin{minipage}[b]{0.5\linewidth}
  \centering
  \centerline{\includegraphics[width=7.5cm]{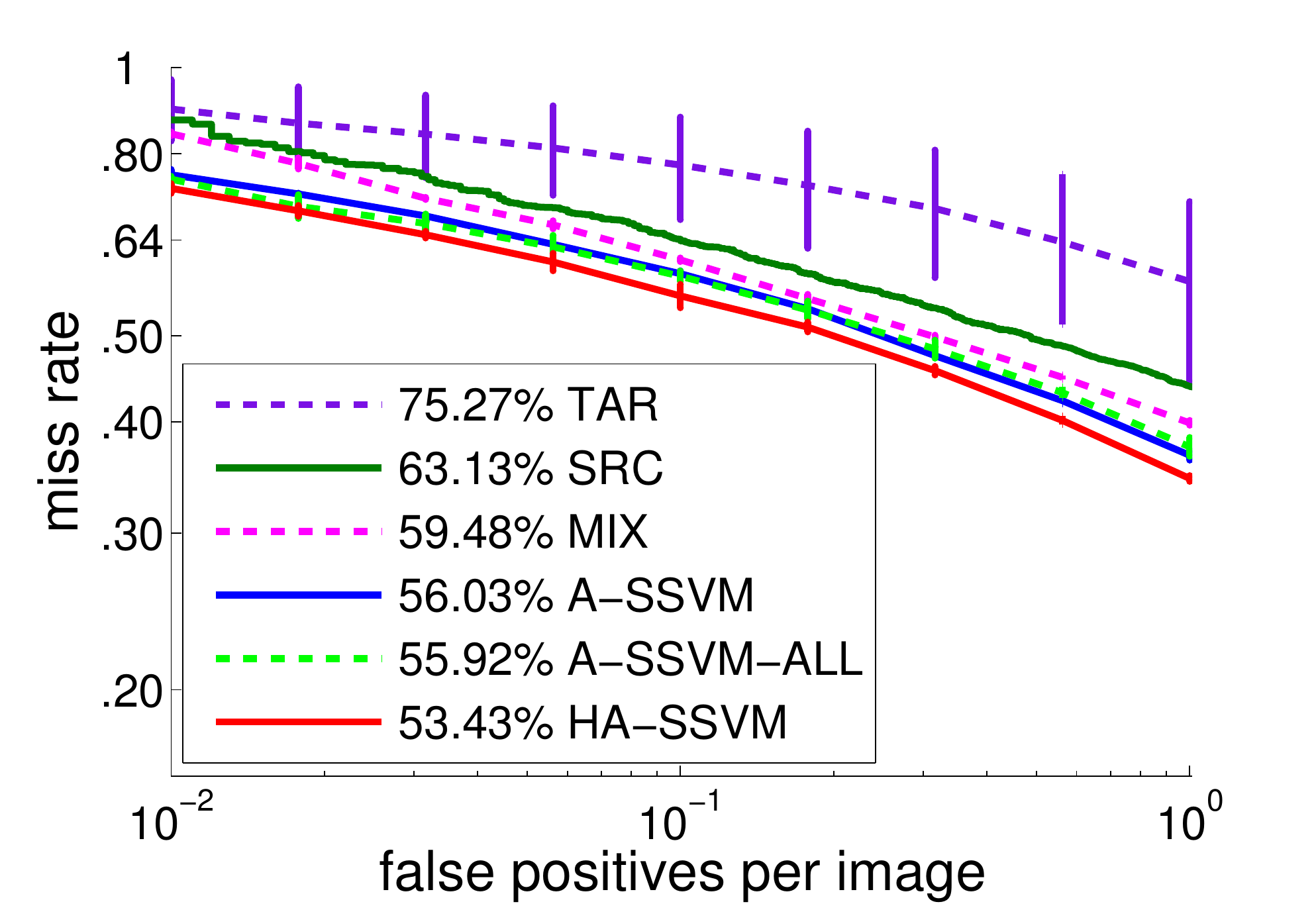}}
  %\vspace{1.5cm}
  \centerline{ETH0}\medskip
\end{minipage}
%\hfill
\begin{minipage}[b]{0.5\linewidth}
  \centering
  \centerline{\includegraphics[width=7.5cm]{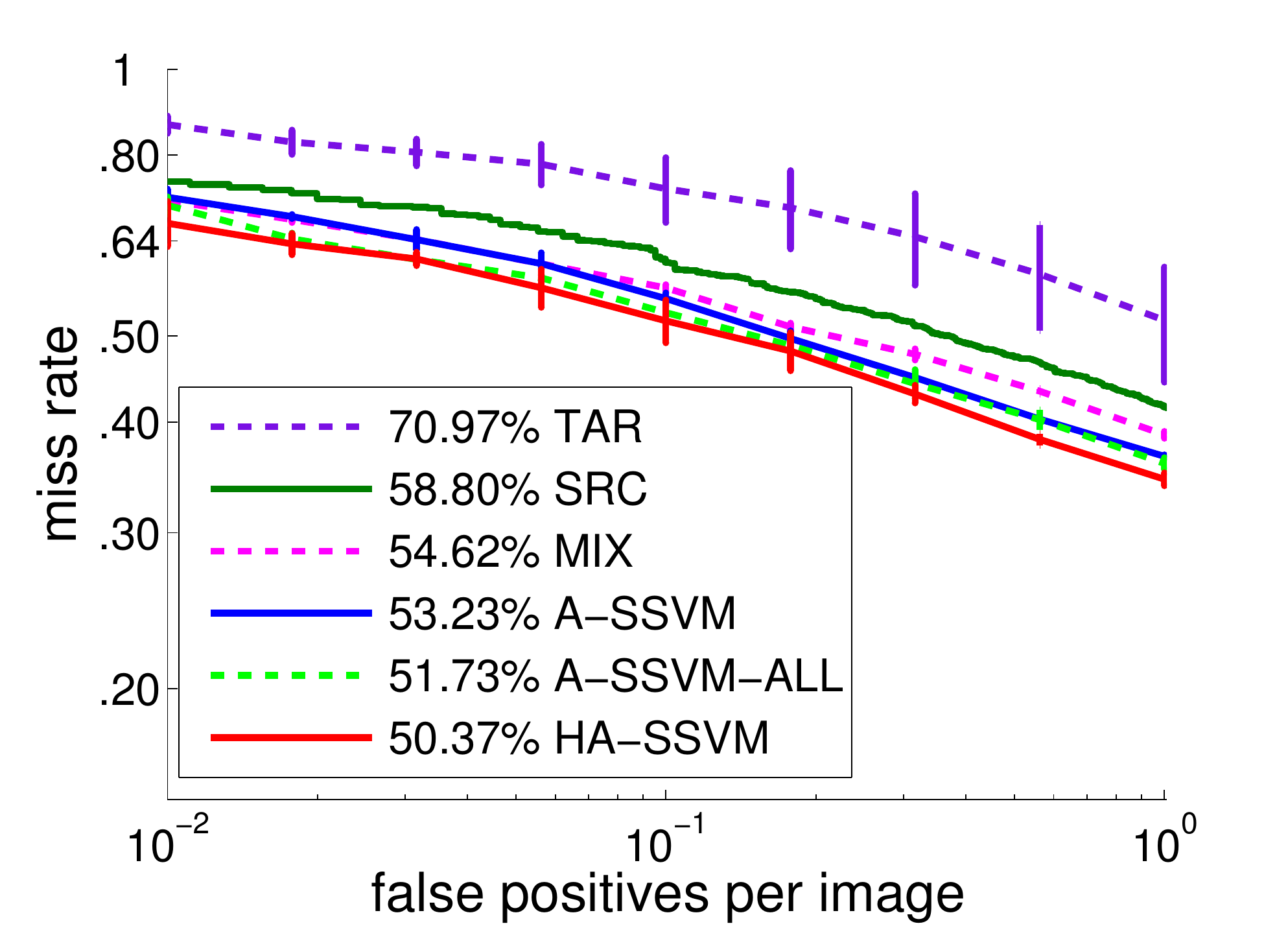}}
  %\vspace{1.5cm}
  \centerline{ETH1}\medskip
\end{minipage}
%\hfill
\begin{minipage}[b]{0.5\linewidth}
  \centering
  \centerline{\includegraphics[width=7.5cm]{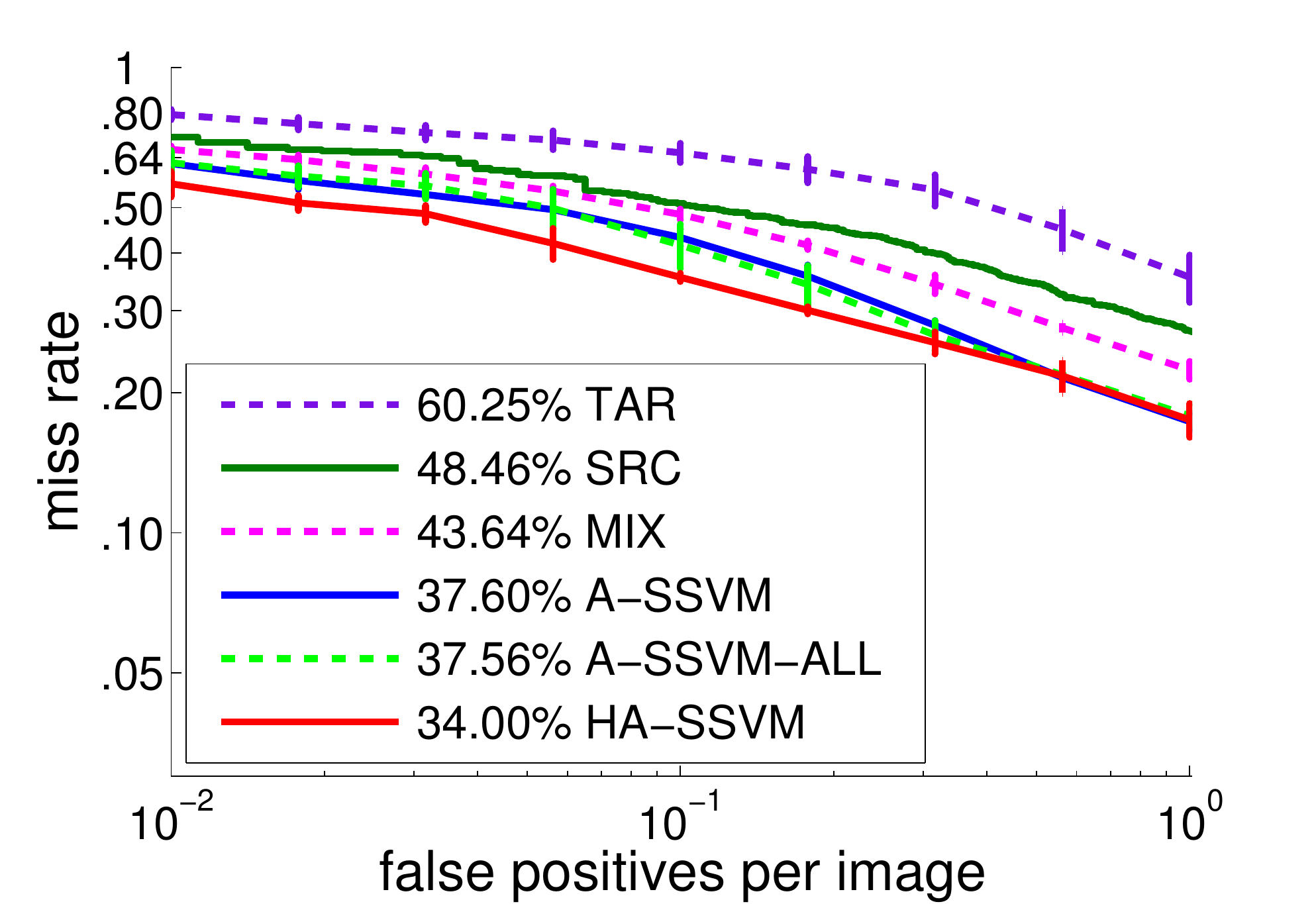}}
  %\vspace{1.5cm}
  \centerline{ETH2}\medskip
\end{minipage}
\begin{minipage}[b]{0.5\linewidth}
  \centering
  \centerline{\includegraphics[width=7.5cm]{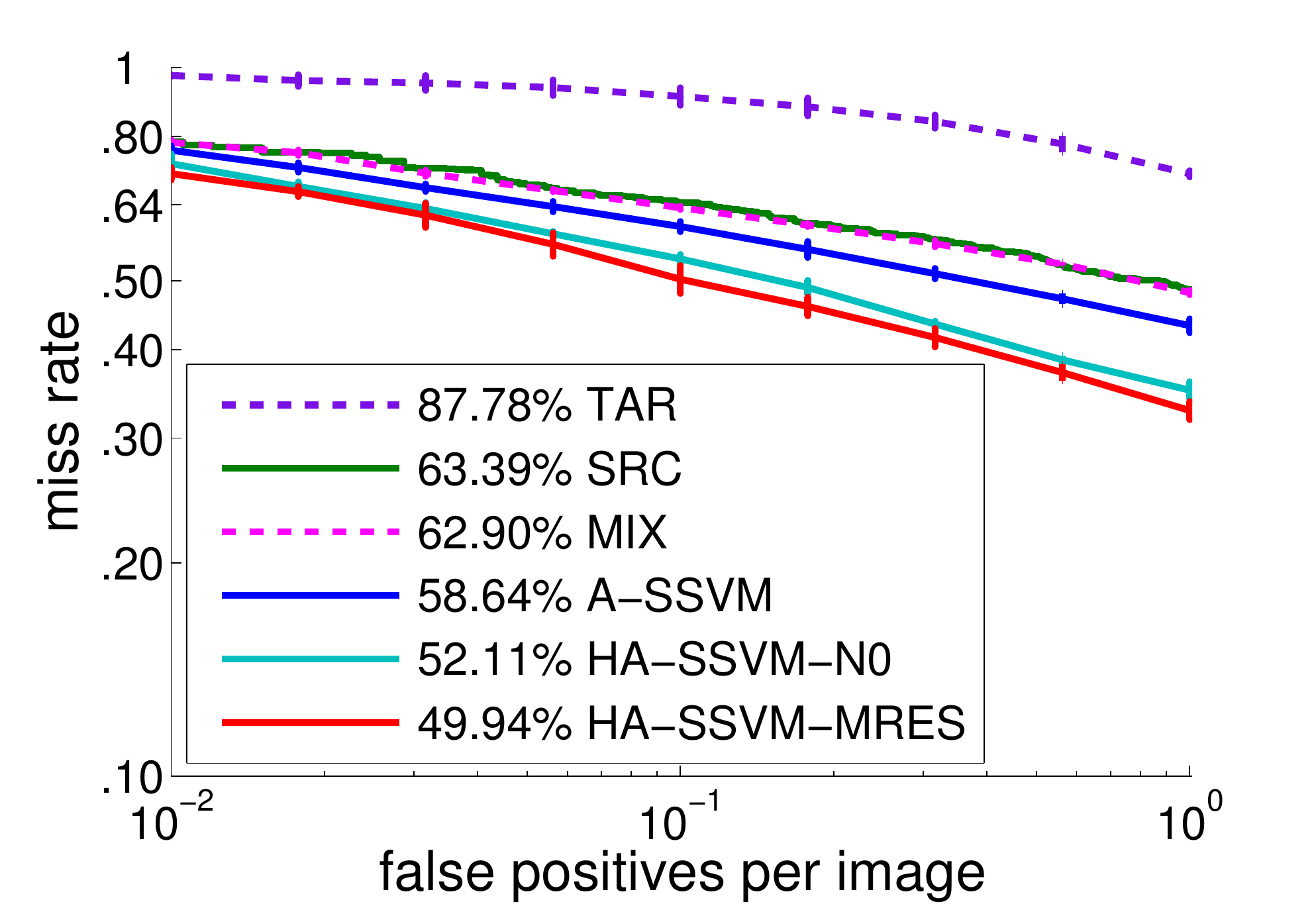}}
  %\vspace{1.5cm}
  \centerline{Caltech}\medskip
\end{minipage}
\caption{ETH0, ETH1 and ETH2 show the adaptation results from virtual-world to ETH three sub-datasets. Caltech shows results of adapting virtual-world DPM detector to a multi-resolution detector in Caltech pedestrian dataset. A-SSVM is trained with mixed high and low resolution samples. HA-SSVM-N0 corresponds to $W^{N_0}$ in the multi-resolution adaptation tree and HA-SSVM-MRES is the adapted multi-resolution detector.}
\label{fig:V-All}
\end{figure*}

\subsubsection{Setup} In supervised domain adaptation it is assumed that there are available just a few labeled data from the different target domains. In order to emulate this setting, we selected only $100$ pedestrians for the experiments with ETH0, ETH1, ETH2 and Caltech, which roughly correspond to the $6\%$, $1.5\%, 3\%, 5\%$, respectively, of the available pedestrians for training. We use all the training pedestrian-free images of these datasets, {\ie}, $999, 451, 354, 1,824$ images, respectively. We follow the Caltech evaluation criterion \citep{Dollar:2012} and plot the average miss rate {\vs} false positive per image (FPPI) curve. We use the suggested reasonable setting and therefore test on the pedestrians taller than $50$ pixels. Each train-test experiment is repeated five times and we report the mean and standard deviation of the repetitions. To ensure fair comparisons, we use the same random samples for different training methods. To evaluate the performance of HA-SSVM, we compare it to the baselines described in Table~\ref{tb_learned_classifiers_dpm}. 

In fact, as touchstone of HA-SSVM for pedestrian detection, our first test was the participation in the {\em Pedestrian Detection Challenge} of the KITTI benchmark\footnote{{\tt http://www.cvlibs.net/datasets/kitti/}} as part of the {\em Reconstruction Meets Recognition Challenge (RMRC)} held in conjunction with the ICCV'2013 celebrated in Sydney. At that time we did not have written neither a report nor this paper, so we participated with the multi-resolution HA-SSVM DPM described here, but with the generic name of {\em DA-DPM} (domain adaptive DPM). In this case, we used $200$ pedestrians of the KITTI training set, roughly the $11\%$ of the available ones, as well as $2,000$ pedestrian-free images of the $7,518$ available for training. We note that under the KITTI benchmark, the object detection evaluation criterion is different from the Caltech one. Accuracy is measured as precision {\vs} recall instead of miss rate {\vs} FPPI. Note also that, in order to avoid parameter tuning, the ground truth of the KITTI testing data is not available.

For all the experiments, the SRC classifier (see Table~\ref{tb_learned_classifiers_dpm}) is the same DPM, trained with the virtual-world dataset. It is worth to mention that we use a DPM root filter of $12 \times 6$ HOG cells (each cell is of $8 \times 8$ pixels), {\ie}, the minimum size of the detectable pedestrians is $96 \times 48$ pixels. Then, for the multi-resolution adaptation (to Caltech and KITTI), we build the two-layer hierarchical model of \figurename~\ref{fig:AdaptationTree}(b). When computing features, we add an extra octave at the bottom of the feature pyramid and then divide the pyramid into two pyramids: high resolution pyramid which contains pedestrians taller than $96$ pixels, and low resolution pyramid which contains pedestrians lower than $96$ pixels. The extended feature pyramid is illustrated in \figurename~\ref{fig:hasvm_multires}(b). During training time, we assign the training pedestrians to the high and low resolution domains according to the height of their bounding boxes, while the background samples are shared by both domains. In figure \figurename~{\ref{fig:CHis}} we show the pedestrian height distribution of the virtual- and real-world training datasets. Note that the virtual-world dataset has few low resolution pedestrians compared with the real-world ones, thus making the pursued adaptation challenging. In testing time, we apply the two adapted models to the corresponding resolution pyramid and finally we combine their detections and apply non-maximum suppression to obtain the final detections. 

\begin{figure*}
\begin{minipage}[b]{1.0\linewidth}
  \centering
  \centerline{\includegraphics[width=17.5cm]{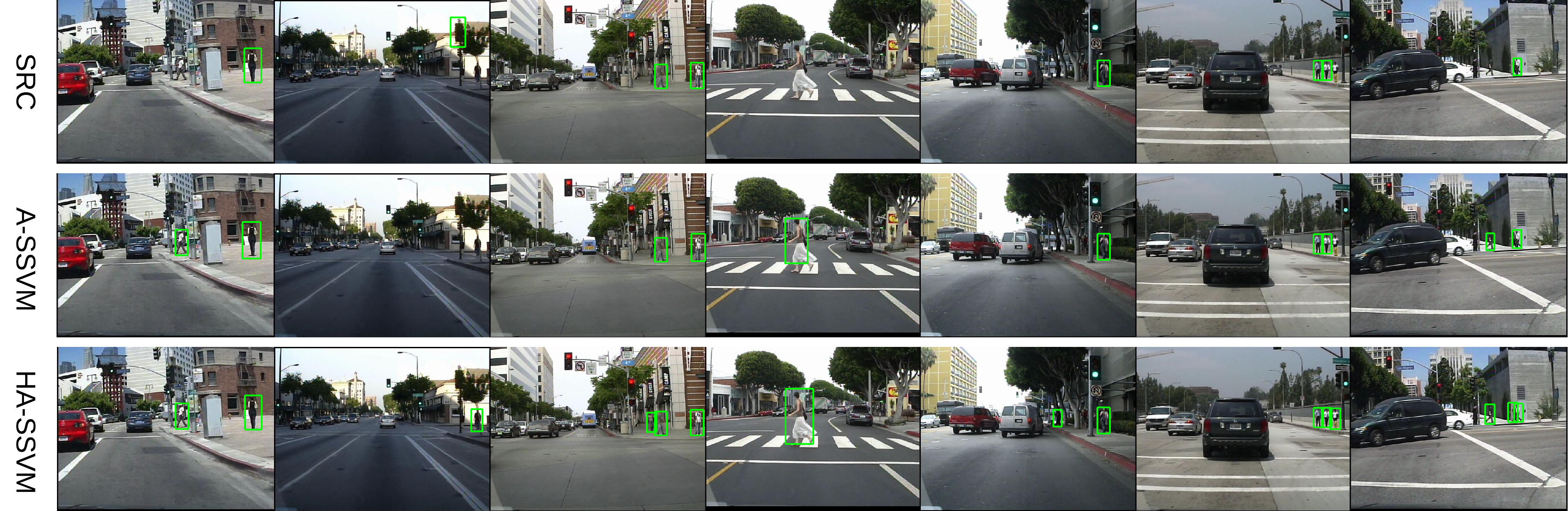}}
  %\vspace{1.5cm}
  \centerline{}\medskip
\end{minipage}
\caption{Pedestrian detection on Caltech dataset for SRC, A-SSVM and HA-SSVM based models. The results are drawn at FPPI = 0.1.}
\label{fig:detections_caltech}
\end{figure*}
%-------------------------------------------------------------------------------
\begin{table}
\begin{center}
\input{table_kit.tex}
\end{center}
\caption{Evaluation on KITTI pedestrian detection benchmark during the RMRC'2013. Results are given as average precision (AP). Our method DA-DPM (which is actually the application of HA-SSVM to a DPM trained with virtual-world data) outperforms the previous best method LSVM-MDPM-sv by $5 \sim 8$ points in average precision. "LP" means that the method uses point clouds from a Velodyne laser scanner. }
\label{tb:kIT}
\end{table}
%------------------------------------------------------------------------------- 

\subsubsection{Results} 
In \figurename~\ref{fig:V-All} we can see the results for the setting of \figurename~\ref{fig:AdaptationTree}(a).  It can be appreciated that pooling-all-target-domains strategy (A-SSVM-ALL) can have better adaptation accuracy than using single target domain data (A-SSVM). However, HA-SSVM achieves even better accuracy when trained with the same samples as A-SSVM-ALL, which demonstrates the importance of leveraging multiple target domains in a hierarchy.

In \figurename~\ref{fig:V-All} we can also see the results of applying setting \figurename~\ref{fig:AdaptationTree}(a) to Caltech. We additionally assessed the accuracy provided by the intermediate model $\textbf{w}^{N0}$, which is denoted by HA-SSVM-N0 in contrast to HA-SSVM-MRes which corresponds to the full multi-resolution adaptation. Note how even HA-SSVM-N0 shows better classification accuracy than A-SSVM. It demonstrates the effectiveness of the progressive adaptation in HA-SSVM, which can be explained by the fact that the multi-task training learns general shared parameters for multiple target domains ({\ie} high- and low-resolution domains), while single-task A-SSVM does not take into account the differences of multi-resolution samples. Of course, HA-SSVM-MRes is providing the best accuracy. Quantitative results are shown in \figurename~{\ref{fig:detections_caltech}}, where it can be seen that HA-SSVM-MRes is capable of detecting lower resolution pedestrians.

\begin{figure}
\begin{minipage}[b]{1.0\linewidth}
  \centering
  \centerline{\includegraphics[width=7cm]{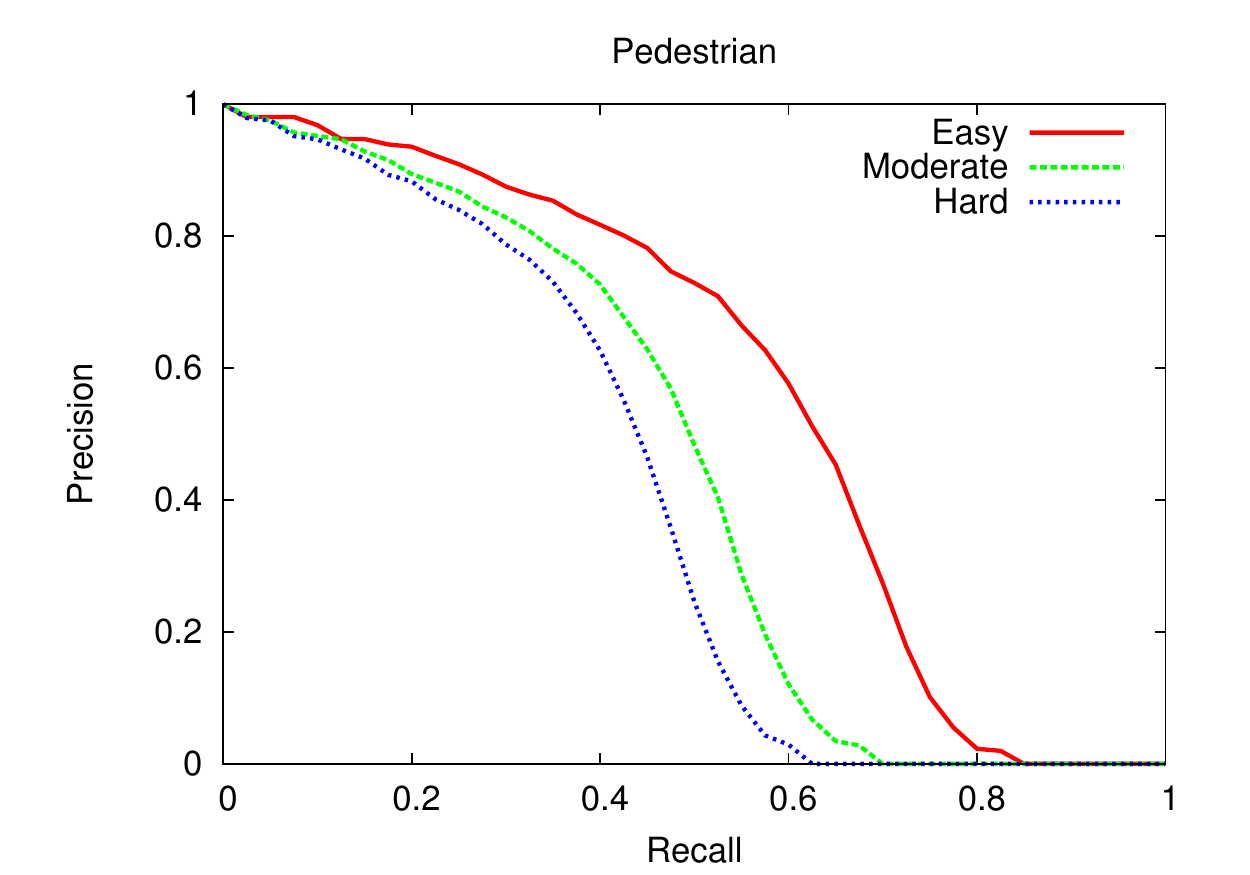}}
  %\vspace{1.5cm}
  % \centerline{Adaptation Tree}\medskip
\end{minipage}
\caption{Pedestrian detection results on KITTI benchmark.}
\label{fig:V-KIT}
\end{figure}

Finally, as can be see in Table \ref{tb:kIT}, we won the pedestrian detection challenge of the RMRC'2013\footnote{In the RMRC'2013 program it can be checked that we did a talk as winners of the pedestrian detection challenge, see {\tt http://ttic.uchicago.edu/~rurtasun/rmrc/program.php}.}, {\ie}, we outperformed LSVM-MDPM-sv \citep{Geiger:2011}, LSVM-MDPM-us \citep{Felzenszwalb:2010} and mBoW \citep{Behley:2013}. Our precision-recall curves can be seen in \figurename~{\ref{fig:V-KIT}}.

We think that this was a quite remarkable result because, as we mentioned before, in order to adapt our virtual-world based pedestrian DPM we only used the $\sim 11\%$ of the KITTI training pedestrians and the $\sim 27\%$ of the available pedes\-trian-free images. This implies that the adaptation took $20$ minutes approximately in our 1 core @ 3.5 Ghz desktop computer, while training the original DPM \citep{Felzenszwalb:2010} with all the full KITTI training set may need around 10 hours in the same conditions. It is also worth to point out that, as can be deduced from \figurename~{\ref{fig:CHis}}, the number of (virtual-world) pedestrians used for building our source model plus the $200$ pedestrians selected from the KITTI training dataset is still lower than the total number of pedestrians in the KITTI dataset. Similarly for the pedestrian-free images, since we use $2,000$ from the virtual world to build the source model and $2,000$ from the KITTI training set to do the adaptation, while there are around $7,500$ available for training. Moreover, as to the best of our knowledge, this was the first time that a domain adapted object detector wins such a challenge.

Of course, after RMRC'2013 we were completing the work presented in this paper, and other proposals have been submitted to the pedestrian detection challenge\footnote{{\tt http://www.cvlibs.net/datasets/kitti/eval\_object.php}}. In particular, at this moment there are six new submissions. Some performed worse than ours (three), some performed quite similarly (two) and another clearly outperformed the rest. Altogether, our domain adapted detector still ranks fourth. We plan to improve our detector in terms of used features and pedestrian model, however, this is out of the scope of this paper since these improvements are not related to the domain adaptation process itself.

%=============================================================================================================================
\section{Domain Adaptation of Multi-category Classifiers}
\label{sec:DA_ObjectRecognition}

In the following, we evaluate HA-SSVM on multi-category classifiers. We begin with the scenario in which the target sub-domains are given a priori. After we assess the scenario in which such sub-domains must be discovered. For illustrating how HA-SSVM operates with multi-category classifiers, we focus on object category recognition.
%Thus, before applying HA-SSVM, we use a domain discovery approach to construct the hierarchical structure of the target domain. 
%\footnote {We will make the source code available.}.
%\label{subsec:SSVM}

%Now we consider the case where the target domain labels are available. 
Assume we are given an set of $N$ examples, $\mathcal{D}$, each one labeled as belonging to a category among $K$ possible ones, {\ie} $\mathcal{D} = \{(\textbf{x}_i, y_i)|\textbf{x}_i \in \mathbb{R}^n, y_i \in \{1,\dots,K\}\}_{i=1}^N$. Let $\textbf{w}_1, \dots, \textbf{w}_K$ be the parameters of $K$ linear category classifiers, so that a new example $\textbf{x}$ is assigned to a category according to the rule  $f(\textbf{x}) = \argmax_{k \in \{1,\dots,K\}} \textbf{w}'_k\textbf{x}$. Let $\textbf{w} = [\textbf{w}'_1,\dots,\textbf{w}'_K]'$ in $\mathbb{R}^{Kn}$, and a feature map $\Phi(\textbf{x}, y) = [\textbf{0}',\dots,\textbf{x}',\dots,\textbf{0}']'$, where $\textbf{0} \in \mathbb{R}^{n}$ is the zero vector and  $\textbf{x}$ is located at the $y$-th slot in $\Phi(\textbf{x}, y)$. Now the multi-category classification problem can be treated as a special case of structure output prediction:
\begin{equation}
f(\textbf{x}) = \argmax_{y \in \{1,\dots,K\}} \textbf{w}'\Phi(\textbf{x}, y).
\end{equation}
 In order to apply HA-SSVM, \Eq{eq:ASSVM} can be directly used as a basic adaptation unit by writing the loss term as: 
\begin{equation}
\begin{array}{ll}
\sum_{i=1}^{N} [ \max_{\hat{y} \in \{1,\dots,K\}}(\textbf{w}'\Phi(\textbf{x},\hat{y}) &+ \Delta(\hat{y}, y_i)) - \textbf{w}'\Phi(\textbf{x}_i, y_i)],
\end{array}
\end{equation}
where $\Delta(\hat{y}, y_i)$ is the $0$-$1$ loss function.
%\section{Extensions}
%(1) What if we reach the maximum layer in the hierarchical tree? It means the prior model is finally adapted to each individual target sample. Conceptually, the final models are exemplar models, which is similar to exemplar SVMs of \citep{}.
%
%(2) Currently, the hierarchical structure in the target domain is assume to be given. The more challenging case arises when such structure are not explicitly given. This could be another interesting topic, {\ie} discovering latent or hidden hierarchical structure of the target data. We can leave this as a future work.
%
%(3) Issue (2) is also related to unsupervised learning, where even the label information is not available.  
%
%(4) Consider the situation of incrementally adding target data. How to assign new coming samples to the previously constructed tree is a problem. More over, should we dynamically update the tree structure? Then how to update it to make it better fit target data. 
%====================================================================================================================
\subsection{Known Target Sub-Domains}
\label{sec:Experiments_ObjRec}

\begin{center}
\begin{table}[t]
{\scriptsize \input{table_learned_classifiers_multicategory.tex}}
\caption{Different types of learned multi-category classifiers.}
\label{tb_learned_classifiers_multicategory}
\end{table}
\end{center}
\begin{table*}[!t]
\begin{center}
\input{table1.tex}
\end{center}
\begin{center}
\input{table2.tex}
\end{center}
\caption{Multi-category recognition accuracy on target domains. Bold indicates the best result for each domain split. Underline indicates the second best result. The domains are: A: \emph{amazon}, W: \emph{webcam}, D: \emph{dslr}, C: \emph{Caltech256}. We use the nomenclature {\em Source}$\to${\em Target}.}
\label{tb:all_domains}
\end{table*}
\begin{table*}
\begin{center}
\input{table3_h.tex}
\end{center}
\caption{Multi-category recognition accuracy averaged across all domain splits, with the corresponding standard deviation.}
\label{tb:avg_domains}
\end{table*}

%\subsubsection{Multi-category Object Recognition}
\subsubsection{Datasets} We evaluate HA-SSVM for object category recognition using the benchmark domain adaptation dataset known as \emph{Office-Caltech} \citep{Gong:2012, Hoffman:2013}. This dataset combines the \emph{Office} \citep{Saenko:2010} and \emph{Caltech256} \citep{Griffin:2007} datasets. In particular, \emph{Office-Caltech} consists of the 10 overlapping object categories between \emph{Office} and \emph{Caltech256}, which are {\tt backpack}, {\tt calculator}, {\tt coffee-mug}, {\tt computer-keyboard}, {\tt computer-monitor}, {\tt computer-mouse}, {\tt head-phones}, {\tt laptop-101}, {\tt touring\--bike} and {\tt video-projector} in the terminology of \emph{Caltech256}.

From the viewpoint of domain adaptation, \emph{Office-Caltech} consists of four domains. One domain, called \emph{caltech} (C), corresponds to the images of \emph{Caltech256}, which were collected from the internet using Google. The other three domains come from \emph{Office}, namely the \emph{amazon} (A), \emph{webcam} (W) and \emph{dslr} (D) domains. The \emph{amazon} domain is a collection of product images from {\tt amazon.com}. The \emph{webcam} and \emph{dslr} domains contain images taken by a (low resolution) webcam and a (high resolution) digital single-lens reflex camera, respectively. Cross-domain variations are not the only ones, but for a particular domain and category, the objects are imaged under different poses and illumination conditions.

\subsubsection{Setup} \label{sss:setup1} We follow the experimental setup of \citep{Saenko:2010, Gong:2012, Hoffman:2013}, which we summarize in the following. We have four domains (A, W, D, C) and the same 10 object categories per domain. For each experiment, one domain is selected as source domain and the other three as target domains. The number of examples per category varies from domain to domain and from category to category. When A is the source, 20 examples are randomly selected per category for training, while when the sources are either W, D or C, only 8 examples are selected per category. When a domain plays the role of target, only 3 examples are selected per category for performing the domain adaptation (training). All the examples of the target domains not used for training are used for testing. The accuracy of the classification is measured as the number of correctly classified test examples divided by the total number of them ({\ie}, without distinguishing object categories). In fact, since the splitting of the available examples into training (source and target) and testing (target) is based on random selection, each experiment is repeated 20 times. Therefore, the average of the 20 obtained accuracy values is actually used as final accuracy measure together with its associated standard deviation. 

In order to make easier across-paper comparisons, we use the same 20 random train/test splits available from \citep{Hoffman:2013}. Moreover, rather than using our own feature computation software, we use the pre-computed SURF-based bag of (visual) words (BoW) available for the images of \emph{Office-Caltech}. Then, following \citep{Gong:2012}, we apply PCA to such original SURF-BoW to obtain histograms of 20 visual words (bins).

\subsubsection{Baselines} We compare our algorithm to the baselines summarized in Table \ref{tb_learned_classifiers_multicategory}. A-SVM, PMT-SVM, A-SSVM and A-SSVM-ALL are adapted with the target domain examples and the source classifiers, the rest of methods require the target domain examples and the original source domain ones for retraining. For the A-SVM and PMT-SVM methods we use the implementation provided by \citep{Aytar:2011}, including MOSEK optimization \citep{Mosek:2013}. We run GFK and MMDT using the code of \citep{Hoffman:2013}. Note that in \citep{Hoffman:2013}, GFK and MMDT are the best performing methods among others, including ARCT \citep{Kulis:2011} and HFA \citep{Duan:2012} methods. All these methods, except A-SSVM-ALL, follow the one-to-one domain adaptation style (\fig{fig:Model}(a)), {\ie}, an independent domain adapation is performed for each target domain. 

\begin{table}
\begin{center}
\input{table_3layer.tex}
\end{center}
\caption{HA-SSVM trained with various adaptation trees. The first column illustrates the tree structure. A two-layer adaptation tree is represented by X$\to$[Y,Z,T], where X is the source domain and Y, Z and T are sibling target domains. These results are just a copy of the HA-SSVM ones shown in Table~\ref{tb:all_domains}. A three-layer adaptation tree is represented by X$\to$[Y,[Z,T]], where Z and T are siblings on the third layer and Y is located on the second layer.}
\label{tb:3 layers}
\end{table}
\begin{table}
\begin{center}
\input{table_QDS.tex}
\end{center}
\caption{Domain similarities in terms of QDS values \citep{Ni:2013}. Lager values indicate higher similarity.}
\label{tb:QDS}
\end{table}

\subsubsection{Results} We first evaluate the accuracy of HA-SSVM with a two-layer adaptation tree, {\ie}, all the target domain datasets are at the same layer and connected to the source domain dataset by an intermediate node, similar to \figurename~\ref{fig:AdaptationTree}(a). The accuracy for each source/targets split is shown in Table~\ref{tb:all_domains}. Table \ref{tb:avg_domains} shows the accuracy of each algorithm averaged over all domain splits. It is worth to note that our results for GFK and MMDT are totally in agreement with the ones presented in \citep{Hoffman:2013} for the same experiments and settings.

From Table~\ref{tb:all_domains} and Table \ref{tb:avg_domains}, it is clear the importance of using all the available target-domain examples. Note that the best performing methods, A-SSVM-ALL (\fig{fig:Model}(b) style) and HA-SSVM (\fig{fig:Model}(c) style), do so in contrast to the rest of methods, which follow the one-to-one domain adaptation style (\fig{fig:Model}(a)). For instance, if we focus in the A$\to$[W,D,C] case, both A-SSVM-ALL and HA-SSVM use 90 target domain examples simultaneously, {\ie} 3 target domains $\times$ 10 object categories per target domain $\times$ 3 examples per category. 1-to-1 domain adaptation style methods use 30 W examples for performing the A$\to$W domain adaptation, and analogously for A$\to$D and A$\to$C. Therefore, potential commonalities between W, D, and C domains are not used. Moreover, HA-SSVM outperforms A-SSVM-ALL, in agreement with our hypothesis that using the underlying hierarchical structure of the target domains is better than just mixing them blindly.

Focusing then on HA-SSVM, it is also interesting to see if other target domain structure ({\eg}, a three-layer hierarchy) can improve the domain adaptation accuracy obtained so far. We test HA-SSVM with various three-layer adaptation trees. Table~\ref{tb:3 layers} shows the results. The three-layer adaptation tree achieves results as good as the ones of the two-layer tree and some of them are even better. By further analyzing the domain relationships of the {\emph{Office}} and {\emph{Caltech256}} datasets, we found that there are strong connections to previous studies on domain similarity. In particular, to the rank of domain (ROD) \citep{Gong:2012} and the quantification of domain shift (QDS) \citep{Ni:2013}. We show the domain similarities in Table~\ref{tb:QDS} using QDS measurement. We note that the three-layer hierarchies which yield to best accuracies are those that best capture the underlying domain relationship. For instance, in the first group of Table~\ref{tb:3 layers}, A$\to$[C, [D, W]] achieves better accuracy than other adaptation trees, which is in agreement with the fact that [D, W] show higher similarity than [D, C] and [C, W] (see Table~\ref{tb:QDS}).

\subsection{Latent Target Sub-Domains}
Now we consider the scenario where the domain labels are not given a priori for the target data. In particular, we use again the \emph{Office-Caltech} dataset with the same settings than in \sect{sss:setup1}. However, we mix the target datasets by removing the domain labels. In these experiments, we first compare two recent domain discovery algorithms, in particular, latent domain discovery \citep{Hoffman:2012} (we call it {\em LatDD}), and domain reshaping \citep{Gong:2013} (we call it {\em Reshape}). Finally, we evaluate the adaptation accuracies with the discovered domains, using HA-SSVM.

%----------------------------------------------------------------------------------------------------------------------------------------------------------------------
%
\begin{figure*}
\begin{minipage}[b]{0.5\linewidth}
  \centering
  \centerline{\includegraphics[width=9cm]{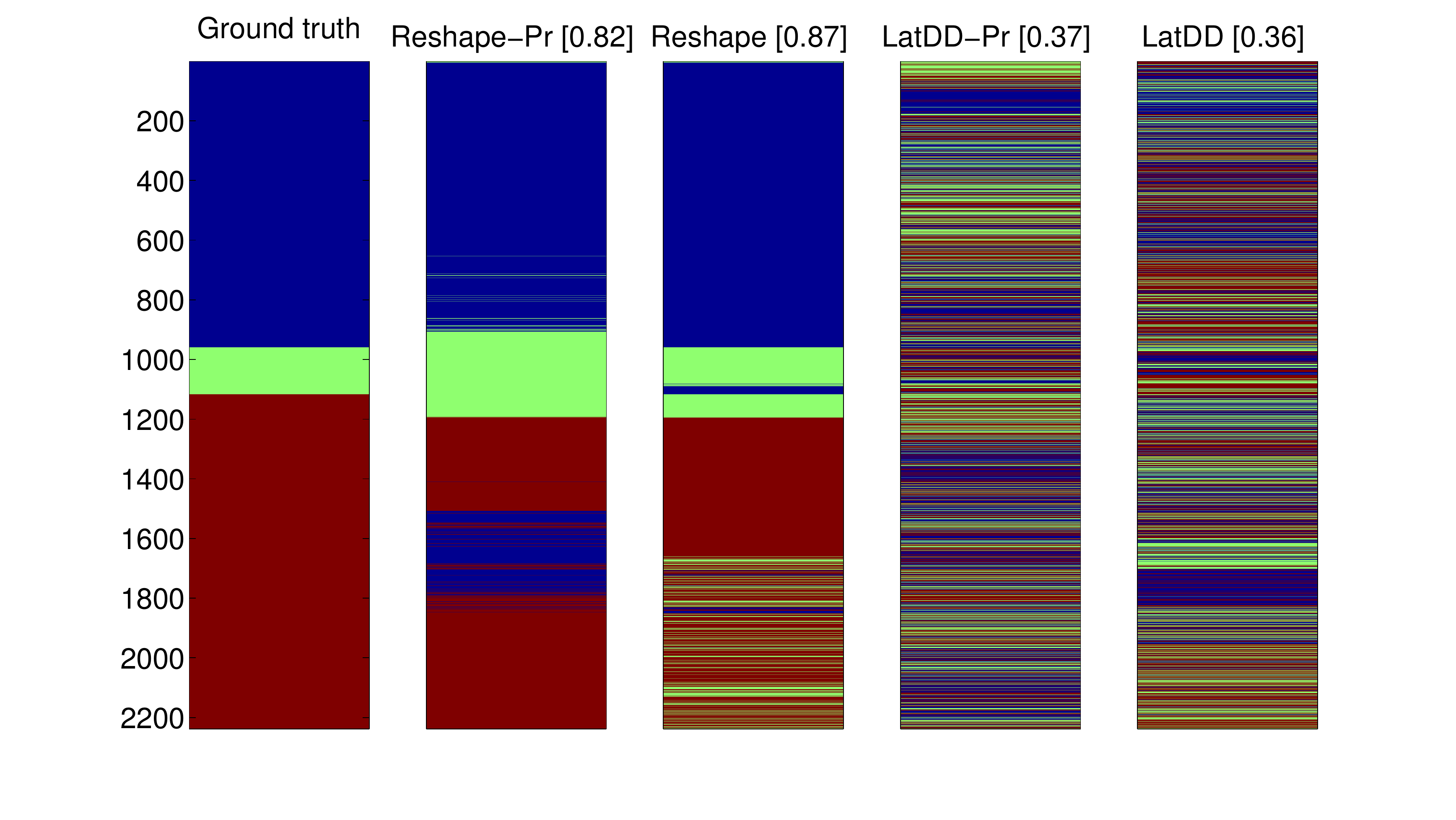}}
  %\vspace{1.5cm}
  \centerline{[A,D,C]}\medskip
\end{minipage}
%\hfill
\begin{minipage}[b]{0.5\linewidth}
  \centering
  \centerline{\includegraphics[width=9cm]{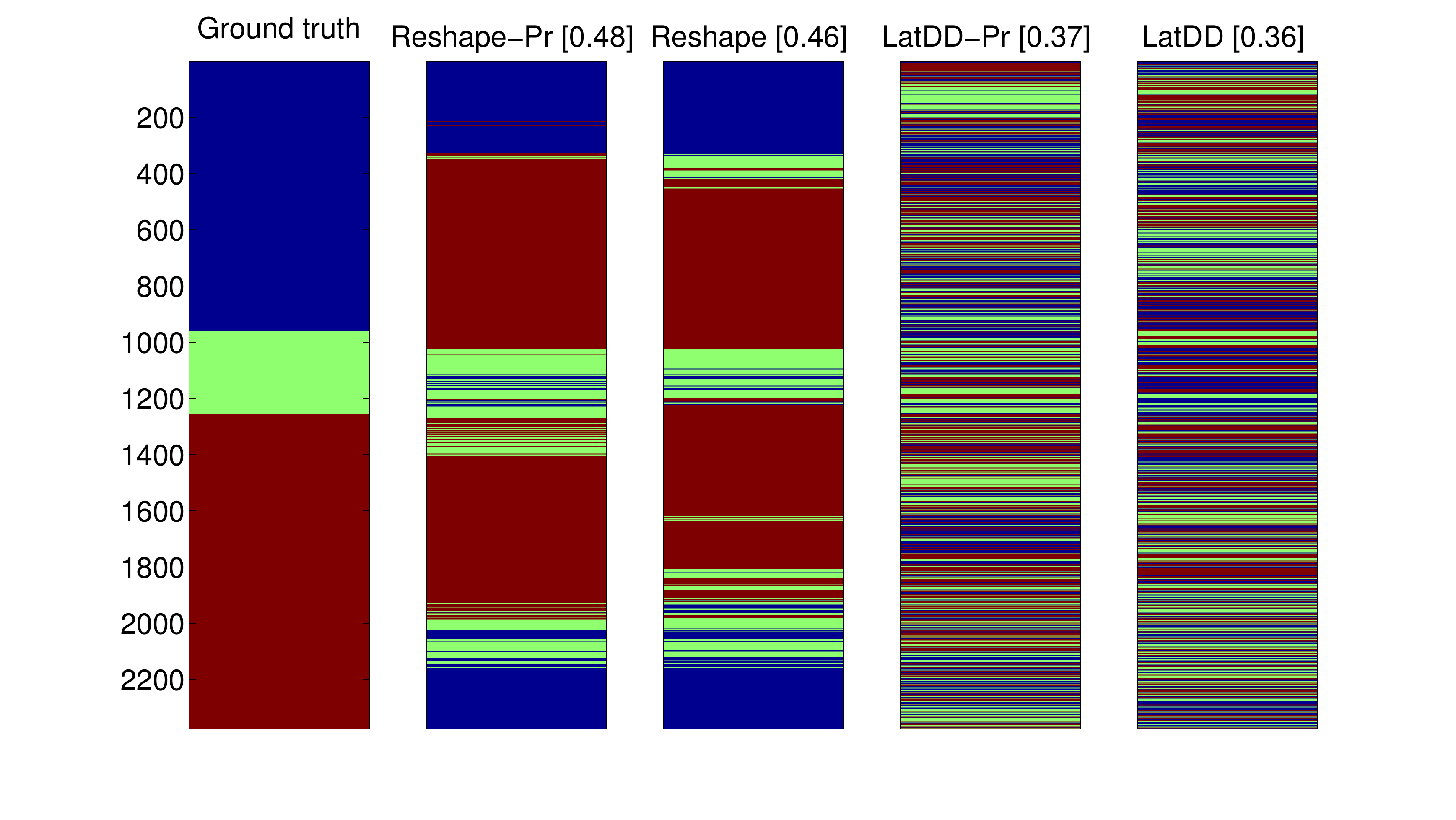}}
  %\vspace{1.5cm}
  \centerline{[A,W,C]}\medskip
\end{minipage}
%\hfill
\begin{minipage}[b]{0.5\linewidth}
  \centering
  \centerline{\includegraphics[width=9cm]{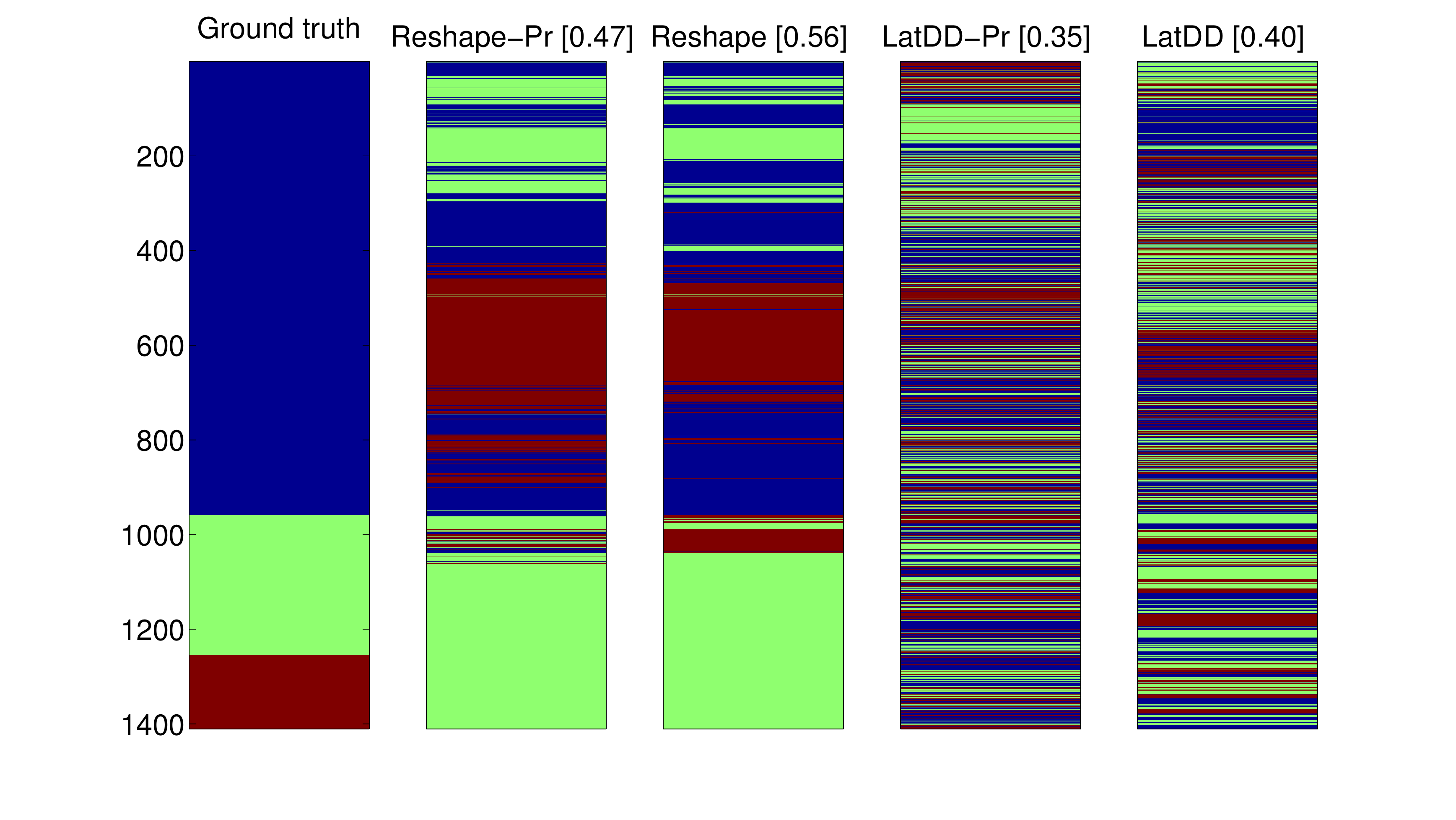}}
  %\vspace{1.5cm}
  \centerline{[A,W,D]}\medskip
\end{minipage}
%\hfill
\begin{minipage}[b]{0.5\linewidth}
  \centering
  \centerline{\includegraphics[width=9cm]{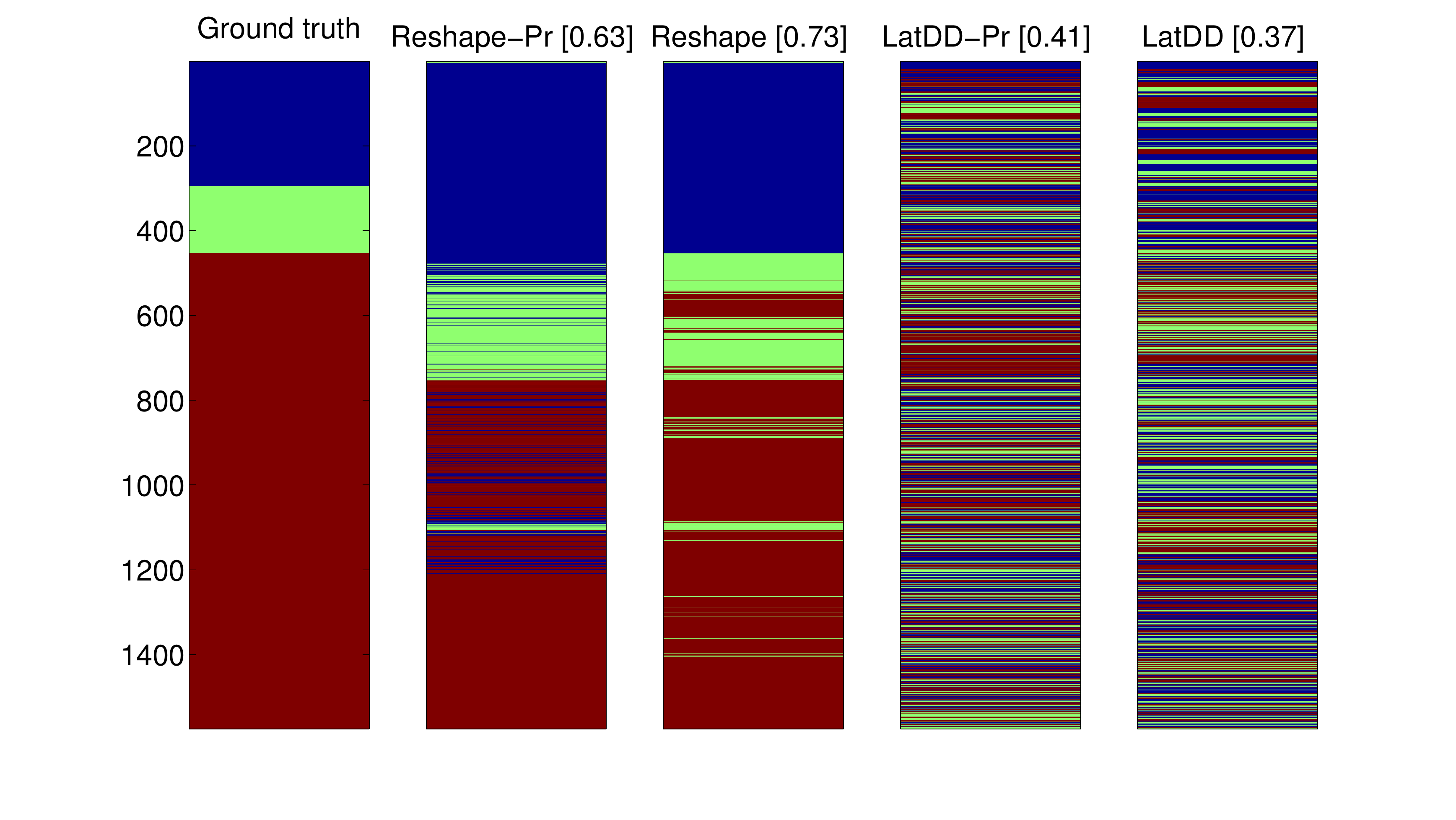}}
  %\vspace{1.5cm}
  \centerline{[W,D,C]}\medskip
\end{minipage}
\caption{Visualization of domain discovery results. The vertical axis indicates the indexes of the examples. Each color represents a different domain. For each sub-figure, the first column shows the original domain (groundtruth). The following columns are domain reshaping with predicted category labels ('Reshape-Pr'), domain reshaping with groundtruth category labels ('Reshape'), latent domain discovery with predicted category labels ('LatDD-Pr') and latent domain discovery with groundtruth category labels ('LatDD'). Within the brackets we show the estimated domain discovery accuracy running in [0,1].}
\label{fig:DomainDiscovery}
\end{figure*}
\begin{figure*}
\begin{minipage}[b]{1.0\linewidth}
  \centering
  \centerline{\includegraphics[width=15cm]{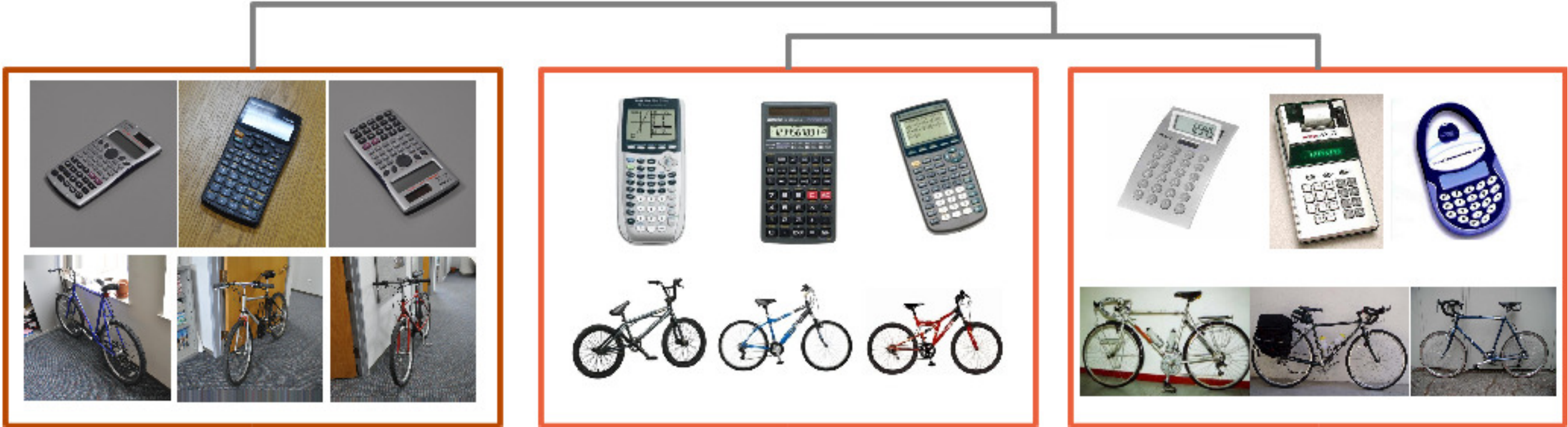}}
  %\vspace{1.5cm}
  \centerline{Original target domains: [A,D,C]}\medskip
\end{minipage} 
\begin{minipage}[b]{1.0\linewidth}
  \centering
  \centerline{\includegraphics[width=15cm]{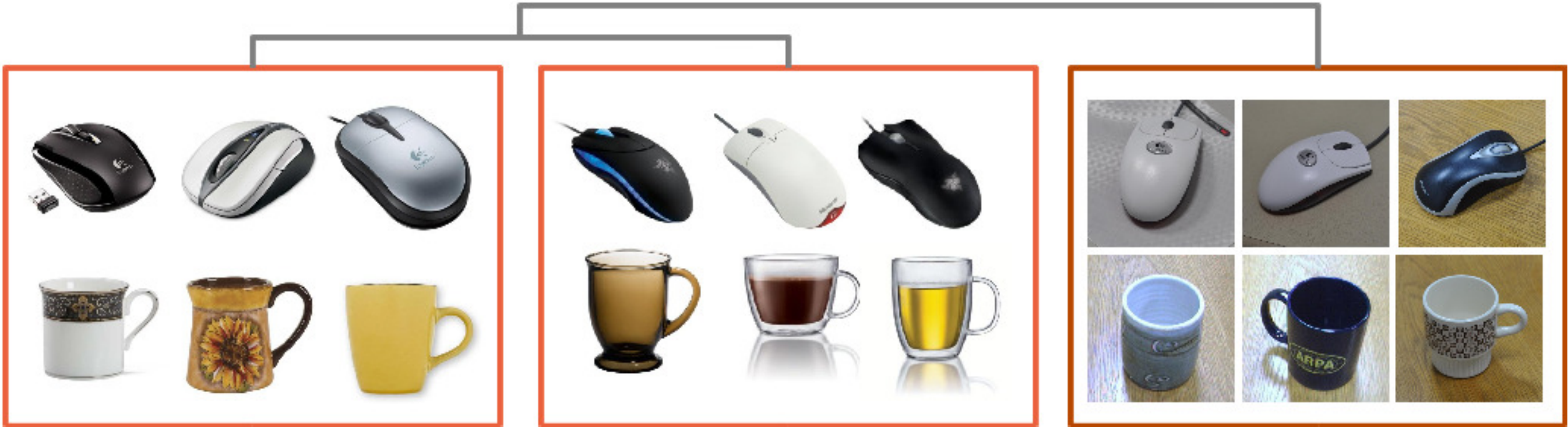}}
  %\vspace{1.5cm}
  \centerline{Original target domains: [A,W,D]}\medskip
\end{minipage}
\caption{{\em Reshape-Pr} + HA-SSVM qualitative results. Three exemplars for two categories are shown for each domain discovered by {\em Reshape-Pr}. The three-layer hierarchy used by HA-SSVM is also indicated for the underlying domains [A,D,C] (top) and [A,W,D] (bottom). In both cases, it correspond to the most accurate HA-SSVM-based multi-category classifier among the different ones that can be obtained for different three-layer hierarchy configurations.}
\label{fig:ExemplarImages_DiscoveredDomains}
\end{figure*}

%\subsubsection{Results}
{\em LatDD} and {\em Reshape} require category labels to operate. However, in our domain adaptation setting we assume that only a few target domain examples have category label, which may be a handicap for such domain discovery methods. In this point, as proof-of-concept, we assumed that the target domain data does not have category labels. Therefore, we first applied the source domain model to predict the category labels in the unlabeled target domain ({\ie}, the domain obtained by mixing the three domains not used as source). We denote by {\em LatDD-Pr} and {\em Reshape-Pr} the cases where we use predicted category labels instead of the groundtruth category labels. 

{\em LatDD} requires as input the number of sub-domains to be discovered (originally this method has been developed to discover source domains), while {\em Reshape} involves an iterative process to search for the optimum number of sub-domains. We want to compare the HA-SSVM results in terms of discovered sub-domains {\vs} a priori given ones, but only from the point of view of how the target data is distributed among a predefined number of target sub-domains. In other words, in these experiments we do not want the number of domains to be discovered. Therefore, we set this value to 3 for fair comparison with the experiments in \sect{sec:Experiments_ObjRec}. It is worth to note that for {\em Reshape/Reshape-Pr} we only use the so-called {\em distinctiveness} maximization step \citep{Gong:2013}.   

\begin{table*}
\begin{center}
\input{table_latent_domain.tex}
\end{center}
\caption{The average multi-category recognition accuracy for a single target domain, for a priori 'Given' domains (\sect{sec:Experiments_ObjRec}), and for discovered latent domains ({\em LatDD-Pr} and {\em Reshape-Pr}). For the three-layer hierarchies we only show the best result among all possible three-layer adaptation trees.}
\label{tb:latent_domains}
\end{table*}

\figurename~\ref{fig:DomainDiscovery} depicts the domain discovery results. It can be seen that {\em Reshape} and {\em Reshape-Pr} are clearly more accurate than {\em LatDD} and {\em LatDD-Pr} predicting the domains. Comparing {\em Reshape} and {\em Reshape-Pr}, we see that the former is more accurate as should be expected since it relies on groundtruth data. Comparing {\em LatDD} with {\em LatDD-Pr}, the accuracy differences are smaller than for {\em Reshape} and {\em Reshape-Pr}. 

Now, for applying HA-SSVM, {\em LatDD-Pr} and {\em Reshape-Pr} are treated equally and as follows. For each discovered sub-domain, we assume that 3 examples are category-labeled for each category. Since our experiments are with 10 categories, as in \sect{sec:Experiments_ObjRec}, 90 target-domain examples must be available for performing domain adaptation (training) and the rest are used for testing. Since this train/test split is based on random selection, we repeat each experiment 20 times in order to emulate the setting of \sect{sec:Experiments_ObjRec}. Note that for {\em LatDD-Pr} and {\em Reshape-Pr} this means that we discard the predicted category labels, but we require only 90 examples to be labeled. In fact, {\em LatDD} and {\em Reshape} are not considered for HA-SSVM since these methods would require the category labels of all the target data (we included them in \figurename~\ref{fig:DomainDiscovery} just as reference to compare with their {\em predicted} counterparts). 

Table~\ref{tb:latent_domains} shows the final domain adaptation accuracies. As in \sect{sec:Experiments_ObjRec} we evaluate two- and three-layer hierarchies, for the latter we only show the best obtained result among all possible configurations. We see that these results are comparable to the best obtained in \sect{sec:Experiments_ObjRec} (also included in Table~\ref{tb:latent_domains} as 'Given' for the reader convenience). Although we work with discovered sub-domains, HA-SSVM still outperforms the single-layer adaptation pooling-all strategy. {\em Resha\-pe-Pr} outperforms {\em LatDD-Pr} as expected given the domain discovery accuracies seen in \figurename~\ref{fig:DomainDiscovery}. However, the differences in accuracy are much larger for domain discovery than for the final object category classification, which may be due to the fact that HA-SSVM trains all the object category classifiers simultaneously for all domains in the hierarchy; thus, partially compensating domain assignment errors. Finally, for illustration purposes, \figurename~\ref{fig:ExemplarImages_DiscoveredDomains} shows object examples within the domains discovered by {\em Reshape-Pr} and some of the three-layer adaptation trees used by HA-SSVM.

\section{Conclusions}
\label{sec:Conclusion}
%\clearpage

In this paper, we present a novel domain adaptation method which leverages multiple target domains (or sub-domains) in a hierarchical adaptation tree. The key idea of the method is to exploit the commonalities and differences of the jointly considered target domains. Given the increasing interest on structural SVM (SSVM) classifiers, we have applied this idea to the domain adaptation method known as adaptive SSVM (A-SSVM), which only requires the target domain samples together with the existing source-domain classifier for performing the desired adaptation. Thus, in contrast with many other methods, the source domain samples are not required. Altogether, we term the presented domain adaptation technique as hierarchical A-SSVM (HA-SSVM). 
%The latent sub-domains of the target data are not always known a priori and the this data may even come without class/category labels. Accordingly, for such an unsupervised scenario we have shown how HA-SSVM can incorporate domain reshaping for discovering a latent sub-domain hierarchy. 

As proof of concept we have applied HA-SSVM to pedestrian detection and object category recognition applications. The former involved to apply HA-SSVM to the widespread deformable part-based model (DPM) while the latter implied their application to multi-category classifiers. In both cases, we showed how HA-SSVM is effective in improving the detection/recognition accuracy with respect to state-of-the-art strategies that ignore the structure of the target data. Moreover, focusing on the object category recognition application, we have evaluated HA-SSVM assuming that the target domains are discovered, obtaining comparable results to the case in which such domains are known a priori. 

As future work we would like to incorporate some recent advances in domain adaptation within the HA-SSVM framework. In particular, our structure-aware A-SSVM (SA-SSVM) approach \citep{Xu_PAMI:2014}, as well as the cross-domain attribute codes \citep{Fatemeh:2013}.

%Include references using bst and bib files.
%\bibliographystyle{mybstfile} 
%{\small \bibliography{mybibfile}}
%\bibliographystyle{abbrv}
%\bibliography{../../../Common/Latex_share/thesis.bib}

%\newpage

%\section{Supplementary Material}
%The complete experiment results on the \textbf{Office + Caltech256 Dataset} are shown in Table \ref{table:full_1}, \ref{table:full_2} and \ref{table:full_3}.
%\begin{table}
%\label{table:full_1}
%\begin{center}
%\input{table_full_1.tex}
%\end{center}
%\caption{Table1}
%\end{table}
%
%\begin{table}
%\label{table:full_2}
%\begin{center}
%\input{table_full_2.tex}
%\end{center}
%\caption{Table2}
%\end{table}
%
%\begin{table}
%\label{table:full_3}
%\begin{center}
%\input{table_full_3.tex}
%\end{center}
%\caption{Table3}
%\end{table}

\begin{acknowledgements}
This work is supported by the Spanish MICINN projects TRA2011-29454-C03-01 and TIN2011-29494-C03-02, Jiaolong Xu's Chinese Scholarship Council (CSC) grant No.2011611023 and Sebastian Ramos' FPI Grant BES-2012-058280.
\end{acknowledgements}

% BibTeX users please use one of
%\bibliographystyle{spbasic}      % basic style, author-year citations
%\bibliographystyle{spmpsci}      % mathematics and physical sciences
%\bibliographystyle{spphys}       % APS-like style for physics
%\bibliographystyle{aps-nameyear}      % American Physical Society (APS) style, author-year citations
\bibliographystyle{apalike}
\bibliography{references}
\end{document}

%% file: shortcuts.tex
\newcommand{\ie}{{\em i.e.}}
\newcommand{\eg}{{\em e.g.}}

\newcommand{\vs}{{\em vs}}
\newcommand{\fig}[1]{Fig. \ref{#1}}

\newcommand{\Eq}[1]{Eq. (\ref{#1})}
\newcommand{\sect}[1]{Sect. \ref{#1}}
\newcommand{\Sect}[1]{Section \ref{#1}}

\usepackage{color}
\definecolor{red}{rgb}{1.00,0.20,0.20}
\definecolor{blue}{rgb}{0.20,0.20,1.00}
\definecolor{green}{rgb}{0.00,1.00,0.00}
\newcommand{\argmax}{\operatornamewithlimits{argmax}}

\usepackage{array}
\newcolumntype{L}[1]{>{\raggedright\let\newline\\\arraybackslash\hspace{0pt}}m{#1}}
\newcolumntype{C}[1]{>{\centering\let\newline\\\arraybackslash\hspace{0pt}}m{#1}}
\newcolumntype{R}[1]{>{\raggedleft\let\newline\\\arraybackslash\hspace{0pt}}m{#1}}

%% file: acronyms.tex
\ifdefined\UseShortAcronyms
\else

\fi

%% file: table_learned_classifiers_dpm.tex
\begin{tabular}{|L{0.15\linewidth}|L{0.7\linewidth}|}\hline
  SRC           &Classifier trained with only the source (virtual-world) domain samples. \\ \hline
  TAR           &Classifier trained with only a relatively low number of target domain (real-world) samples. \\ \hline
  MIX           &Classifier trained with source samples used for SRC and the target domain samples used for TAR. \\ \hline
  A-SSVM        &Classifier adapted with A-SSVM using the SRC model and the target domain samples used for TAR.\\ \hline
  A-SSVM-ALL    &As before but pooling all the considered target domains.\\ \hline
\end{tabular}

%% file: table_kit.tex
%\begin{tabular}{|c | c | c | c | c | c |} \hline
%{ Rank} & { Method} & { Setting} & { Moderate} & { Easy} & { Hard} \\ \hline
%1 & DA-DPM &\textemdash& \textbf{45.51 \%} & \textbf{56.36 \%} & \textbf{41.08 \%}\\
%2 & LSVM-MDPM-sv &\textemdash& 39.36 \% & 47.74 \% & 35.95 \% \\
%3 & LSVM-MDPM-us &\textemdash& 38.35 \% & 45.50 \% & 34.78 \% \\
%4 & mBoW & Laser Points & 31.37 \% & 44.28 \% & 30.62 \% \\ \hline
%\end{tabular}

\begin{tabular}{|c | c | c | c | c |} \hline
{ Rank}  & { Method}    & { Moderate} & { Easy} & { Hard} \\ \hline
1        & DA-DPM       &\textbf{45.51 \%} & \textbf{56.36 \%} & \textbf{41.08 \%}\\
2        & LSVM-MDPM-sv &        39.36 \%  &         47.74 \%  &         35.95 \% \\
3        & LSVM-MDPM-us &        38.35 \%  &         45.50 \%  &         34.78 \% \\
4        & mBoW (LP)    &        31.37 \%  &         44.28 \%  &         30.62 \% \\ \hline
\end{tabular}

%% file: table_learned_classifiers_multicategory.tex
%\begin{tabular}{|L{0.15\linewidth}|L{0.7\linewidth}|}\hline
%ASVM &Adaptive SVM \citep{Yang:2007}. \\ \hline
%
%PMT-SVM &Projective model transfer SVM \citep{Aytar:2011}, which is a variant of ASVM. \\ \hline
%
%ARCT &A general feature transform method proposed in \citep{Kulis:2011}, using both domain data. We compare with the results available from \citep{Hoffman:2013}.\\ \hline
%
%HFK &A feature transform based method that learns a latent common space between source and target as well as  a common space classifier \citep{Duan:2012}. We compare with the results available from \citep{Hoffman:2013}.\\ \hline
%
%GFK &The geodesic flow kernel method \citep{Gong:2012}, using all source and target data (including testing data). We apply 1-nearest neighbor classifier with the kernel as in \citep{Hoffman:2013}.\\ \hline
%
%MMDT &Max-margin domain transfer method of \citep{Hoffman:2013}, which learns a mapping from target domain to source domain as well as a discriminative classifier, using the mapped target features and source domain features.\\ \hline
%
%A-SSVM &as in Table~\ref{tb_learned_classifiers_dpm}. \\ \hline
%A-SSVM-ALL &as in Table~\ref{tb_learned_classifiers_dpm}. \\ \hline
%\end{tabular}

\begin{tabular}{|L{0.15\linewidth}|L{0.7\linewidth}|}\hline
ASVM     & Adaptive SVM \citep{Yang:2007}. It does not require the source domain data, only the learned source classifier. In contrast to A-SSVM, ASVM does not consider structural information. \\ \hline
PMT-SVM  & Projective model transfer SVM \citep{Aytar:2011}, which is a variant of ASVM. \\ \hline
%ARCT     & A general feature transform method proposed in \citep{Kulis:2011}, which requires both the source and target domain data. \\ \hline
%HFA      & A feature transform based method that learns a latent common space between source and target domains as well as a common space classifier \citep{Duan:2012}. \\ \hline
GFK      & The geodesic flow kernel method \citep{Gong:2012}, which requires both source and target domain data (including testing data). \\ \hline
MMDT     & Max-margin domain transfer method of \citep{Hoffman:2013}, which learns a mapping from target domain to source domain as well as a discriminative classifier using the mapped target and source domain features. \\ \hline
A-SSVM   & Analogous to Table~\ref{tb_learned_classifiers_dpm}. \\ \hline
A-SSVM-ALL & Analogous to Table~\ref{tb_learned_classifiers_dpm}. \\ \hline
\end{tabular}

%% file: table1.tex
\begin{tabular}{|r|c|c|c|c|c|c|}\hline 
   & A $\to$ W & A $\to$ D & A $\to$ C & W $\to$ A & W $\to$ D & W $\to$ C\\ \hline \hline
%SRC (Liblinear) & 39.7 {$\pm$} 1.4 & 37.7 {$\pm$} 1.0 & 38.4 {$\pm$} 0.5 & 32.6 {$\pm$} 1.0 & 64.2 {$\pm$} 0.9 & 26.8 {$\pm$} 0.6\\ \hline 
%SRC (Pegasos) & 37.1 {$\pm$} 1.3 & 34.5 {$\pm$} 1.1 & 36.6 {$\pm$} 0.3 & 35.9 {$\pm$} 0.8 & 63.8 {$\pm$} 1.2 & 28.4 {$\pm$} 0.6\\ \hline 
%SRC (QP-Mosek) & 36.9 {$\pm$} 1.3 & 36.7 {$\pm$} 1.1 & 37.9 {$\pm$} 0.4 & 32.3 {$\pm$} 1.0 & 62.8 {$\pm$} 1.0 & 27.1 {$\pm$} 0.5\\ \hline 
%TAR (Liblinear) & 57.1 {$\pm$} 1.0 & 44.2 {$\pm$} 0.9 & 26.5 {$\pm$} 0.6 & 45.1 {$\pm$} 1.2 & 47.9 {$\pm$} 1.7 & 25.5 {$\pm$} 0.8\\ \hline 
%TAR (Pegasos) & 60.0 {$\pm$} 0.8 & 47.0 {$\pm$} 0.9 & 28.9 {$\pm$} 0.8 & 46.3 {$\pm$} 1.0 & 52.9 {$\pm$} 1.6 & 27.0 {$\pm$} 0.7\\ \hline 
%TAR (QP-Mosek) & 64.7 {$\pm$} 1.1 & 51.4 {$\pm$} 1.1 & 30.8 {$\pm$} 0.6 & 48.5 {$\pm$} 1.1 & 54.0 {$\pm$} 1.4 & 29.8 {$\pm$} 1.0\\ \hline \hline

ASVM & 65.0 {$\pm$} 1.0 & 51.6 {$\pm$} 1.1 & 30.9 {$\pm$} 0.6 & 48.6 {$\pm$} 1.1 & 54.4 {$\pm$} 1.5 & 29.8 {$\pm$} 1.0\\ \hline 
PMT-SVM & \underline{65.9} {$\pm$} 1.0 & 52.6 {$\pm$} 1.1 & 32.3 {$\pm$} 0.6 & 49.0 {$\pm$} 1.1 & 57.9 {$\pm$} 1.6 & 30.4 {$\pm$} 0.9\\ \hline \hline

%ARCT          & 55.7 {$\pm$} 0.9          & 50.2 {$\pm$} 0.7          & 37.0 {$\pm$} 0.4          & 43.4 {$\pm$} 0.5          & \underline{71.3} {$\pm$} 0.8 & 31.9 {$\pm$} 0.5          \\ \hline
%HFA           & 61.8 {$\pm$} 1.1          & 52.7 {$\pm$} 0.9          & 31.1 {$\pm$} 0.6          & 45.9 {$\pm$} 0.7          & 57.1 {$\pm$} 1.0 & 29.4 {$\pm$} 0.6          \\ \hline
GFK           & 56.5 {$\pm$} 0.8          & 45.3 {$\pm$} 0.9          & 38.6 {$\pm$} 0.4          & 45.8 {$\pm$} 0.6          & \textbf{73.8} {$\pm$} 0.7 & 32.6 {$\pm$} 0.6          \\ \hline
MMDT          & 65.1 {$\pm$} 1.2          & 54.5 {$\pm$} 1.0          & 39.7 {$\pm$} 0.5          & \underline{50.6} {$\pm$} 0.8          & 62.5 {$\pm$} 1.0 & 34.8 {$\pm$} 0.8          \\ \hline \hline

%SRC-SSVM & 42.3 {$\pm$} 0.8 & 38.4 {$\pm$} 0.6 & 39.8 {$\pm$} 0.3 & 39.2 {$\pm$} 0.4 & 65.3 {$\pm$} 0.7 & 33.7 {$\pm$} 0.5\\ \hline 
%TAR-SSVM & 64.5 {$\pm$} 0.9 & 52.1 {$\pm$} 1.1 & 33.5 {$\pm$} 0.8 & \underline{50.5} {$\pm$} 0.7 & 56.0 {$\pm$} 1.0 & 31.3 {$\pm$} 0.8\\ \hline 
%TAR-ALL & \underline{66.5} {$\pm$} 0.7 & \underline{57.9} {$\pm$} 1.0 & 36.3 {$\pm$} 0.4 & \underline{51.3} {$\pm$} 0.6 & 53.7 {$\pm$} 1.0 & \underline{37.1} {$\pm$} 0.5\\ \hline 
A-SSVM & 60.0 {$\pm$} 0.9 & 49.7 {$\pm$} 0.8 & \textbf{42.6} {$\pm$} 0.5 & 49.5 {$\pm$} 0.5 & 67.4 {$\pm$} 0.7 & 37.3 {$\pm$} 0.5\\ \hline 
A-SSVM-ALL & 64.5 {$\pm$} 0.7 & \underline{55.1} {$\pm$} 0.8 & \textbf{42.6} {$\pm$} 0.3 & 49.8 {$\pm$} 0.5 & \underline{67.8} {$\pm$} 0.8 & \underline{39.0} {$\pm$} 0.3 \\ \hline 
HA-SSVM & \textbf{69.8} {$\pm$} 0.7 & \textbf{59.7} {$\pm$} 0.9 & \underline{42.1} {$\pm$} 0.4 & \textbf{54.4} {$\pm$} 0.6 & 66.1 {$\pm$} 1.1 & \textbf{39.4} {$\pm$} 0.3\\ \hline 
\end{tabular}

%% file: table2.tex
\begin{tabular}{|r|c|c|c|c|c|c|}\hline 
   & D $\to$ A & D $\to$ W & D $\to$ C & C $\to$ A & C $\to$ W & C $\to$ D\\ \hline \hline 
%SRC (Liblinear) & 32.2 {$\pm$} 0.7 & 70.8 {$\pm$} 1.1 & 25.2 {$\pm$} 0.4 & 37.0 {$\pm$} 0.8 & 30.8 {$\pm$} 1.5 & 32.0 {$\pm$} 1.0\\ \hline 
%SRC (Pegasos) & 34.3 {$\pm$} 0.6 & 71.3 {$\pm$} 0.7 & 27.9 {$\pm$} 0.3 & 38.9 {$\pm$} 1.0 & 29.9 {$\pm$} 1.3 & 31.7 {$\pm$} 1.3\\ \hline 
%SRC (QP-Mosek) & 31.2 {$\pm$} 0.9 & 69.4 {$\pm$} 1.1 & 25.6 {$\pm$} 0.5 & 38.3 {$\pm$} 0.9 & 30.2 {$\pm$} 1.3 & 30.1 {$\pm$} 1.1\\ \hline 
%TAR (Liblinear) & 43.0 {$\pm$} 1.0 & 56.4 {$\pm$} 0.9 & 26.6 {$\pm$} 0.7 & 44.4 {$\pm$} 1.1 & 56.6 {$\pm$} 1.0 & 44.5 {$\pm$} 1.6\\ \hline 
%TAR (Pegasos) & 44.9 {$\pm$} 0.9 & 60.9 {$\pm$} 1.0 & 28.4 {$\pm$} 0.8 & 45.8 {$\pm$} 1.1 & 58.2 {$\pm$} 1.0 & 50.0 {$\pm$} 1.4\\ \hline 
%TAR (QP-Mosek) & 47.8 {$\pm$} 1.0 & 63.1 {$\pm$} 1.1 & 29.8 {$\pm$} 0.8 & 49.5 {$\pm$} 1.0 & 63.1 {$\pm$} 1.2 & 52.6 {$\pm$} 1.3\\ \hline \hline

ASVM & 48.0 {$\pm$} 1.1 & 63.5 {$\pm$} 1.1 & 29.9 {$\pm$} 0.8 & 49.5 {$\pm$} 1.0 & 63.2 {$\pm$} 1.2 & 52.7 {$\pm$} 1.3\\ \hline 
PMT-SVM & 48.6 {$\pm$} 1.1 & 66.5 {$\pm$} 1.2 & 30.9 {$\pm$} 0.8 & 50.0 {$\pm$} 1.0 & 64.3 {$\pm$} 1.2 & 52.2 {$\pm$} 1.3\\ \hline \hline 

%ARCT     & 42.5 {$\pm$} 0.5          & \underline{78.3} {$\pm$} 0.5          & 33.5 {$\pm$} 0.4          & 44.1 {$\pm$} 0.6          & 55.9 {$\pm$} 1.0          & 50.6 {$\pm$} 0.8          \\ \hline
%HFA      & 45.8 {$\pm$} 0.9          & 62.1 {$\pm$} 0.7          & 31.0 {$\pm$} 0.5          & 45.5 {$\pm$} 0.9          & 60.5 {$\pm$} 0.9          & 51.9 {$\pm$} 1.1          \\ \hline
GFK      & 45.8 {$\pm$} 0.4          & \textbf{80.3} {$\pm$} 0.7 & 33.3 {$\pm$} 0.5          & 46.4 {$\pm$} 0.7          & 61.0 {$\pm$} 1.4          & 52.7 {$\pm$} 1.2          \\ \hline
MMDT     & \underline{50.4} {$\pm$} 0.7          & 74.2 {$\pm$} 0.7          & 35.7 {$\pm$} 0.7          & \underline{51.1} {$\pm$} 0.7          & 62.9 {$\pm$} 1.1          & 53.0 {$\pm$} 1.0          \\ \hline \hline

%SRC-SSVM & 35.4 {$\pm$} 0.5 & 71.5 {$\pm$} 0.7 & 29.8 {$\pm$} 0.2 & 42.2 {$\pm$} 0.7 & 37.0 {$\pm$} 1.2 & 39.5 {$\pm$} 1.1\\ \hline 
%TAR-SSVM & 49.4 {$\pm$} 0.8 & 66.2 {$\pm$} 0.9 & 33.2 {$\pm$} 0.8 & 50.2 {$\pm$} 0.9 & 63.4 {$\pm$} 1.1 & 53.8 {$\pm$} 1.2\\ \hline 
%TAR-ALL & \underline{50.6} {$\pm$} 0.5 & 60.4 {$\pm$} 0.7 & 37.0 {$\pm$} 0.7 & 47.6 {$\pm$} 0.8 & \underline{66.8}}{$\pm$} 0.7 & 57.3 {$\pm$} 1.4\\ \hline 
A-SSVM & 48.6 {$\pm$} 0.5 & \underline{74.6} {$\pm$} 0.6 & 35.5 {$\pm$} 0.5 & \textbf{53.4} {$\pm$} 0.7 & 63.6 {$\pm$} 1.2 & 52.7 {$\pm$} 1.0\\ \hline 
A-SSVM-ALL & 49.0 {$\pm$} 0.5 & 73.2 {$\pm$} 0.7 & \underline{37.8} {$\pm$} 0.4 & 50.6 {$\pm$} 0.6 & \underline{66.2} {$\pm$} 0.6 & \underline{57.5} {$\pm$} 0.9\\ \hline 
HA-SSVM & \textbf{52.6} {$\pm$} 0.5 & 73.0 {$\pm$} 0.5 & \textbf{39.2} {$\pm$} 0.6 & \textbf{53.4} {$\pm$} 0.8 & \textbf{69.6} {$\pm$} 0.7 & \textbf{61.2} {$\pm$} 0.9\\ \hline 
\end{tabular}

%% file: table3_h.tex
%\begin{tabular}{|c|c|c|c|c|c|c|c|c|}\hline 
%ASVM             & PMT-SVM          & ARCT             & HFA              & GFK              & MMDT             & A-SSVM           & A-SSVM-ALL & HA-SSVM\\ \hline \hline 
%48.9 {$\pm$} 1.1 & 48.9 {$\pm$} 1.1 & 49.5 {$\pm$} 0.6 & 47.4 {$\pm$} 0.8 & 51.0 {$\pm$} 0.7 & 52.9 {$\pm$} 0.9 & 52.9 {$\pm$} 0.7 & \underline{\textit{54.4}} {$\pm$} 0.6 &\textbf{56.7} {$\pm$} 0.7\\ \hline 
%%SRC (Liblinear) & 38.9 {$\pm$} 0.9\\ \hline 
%%SRC (Pegasos) & 39.2 {$\pm$} 0.9\\ \hline 
%%SRC (QP-Mosek) & 38.2 {$\pm$} 0.9\\ \hline 
%%TAR (Liblinear) & 43.1 {$\pm$} 1.0\\ \hline 
%%TAR (Pegasos) & 45.9 {$\pm$} 1.0\\ \hline 
%%TAR (QP-Mosek) & 48.8 {$\pm$} 1.1\\ \hline \hline
%\end{tabular}

\begin{tabular}{|c|c|c|c|c|c|c|}\hline 
ASVM             & PMT-SVM          & GFK              & MMDT             & A-SSVM           & A-SSVM-ALL                            & HA-SSVM                  \\ \hline \hline 
48.9 {$\pm$} 1.1 & 48.9 {$\pm$} 1.1 & 51.0 {$\pm$} 0.7 & 52.9 {$\pm$} 0.9 & 52.9 {$\pm$} 0.7 & \underline{54.4} {$\pm$} 0.6 & \textbf{56.7} {$\pm$} 0.7\\ \hline 
\end{tabular}

%% file: table_3layer.tex
\begin{tabular}{|l|c|c|c|c|}\hline 
Adaptation Tree 		&A$\to$W				&A$\to$D				&A$\to$C					&Avg. 			\\ \hline \hline
A$\to$[W, D, C]		&69.8 {$\pm$} 0.7	&59.7 {$\pm$} 0.9	&42.1 {$\pm$} 0.4		&48.4			\\ \hline
A$\to$[W, [D, C]]	&69.8 {$\pm$} 0.7	&59.5 {$\pm$} 1.0	&40.1 {$\pm$} 0.4		&47.7 			\\ \hline
A$\to$[D, [W, C]]	&69.8 {$\pm$} 0.6	&59.1 {$\pm$} 0.8	&40.9 {$\pm$} 0.4		&47.8 			\\ \hline
A$\to$[C, [D, W]]	&72.7 {$\pm$} 0.7	&63.4 {$\pm$} 1.2	&42.1 {$\pm$} 0.4		&\textbf{49.5} 	\\ \hline \hline
Adaptation Tree 		&W$\to$A				&W$\to$D				&W$\to$C 				&Avg. 			\\ \hline \hline
W$\to$[A, D, C]		&54.4 {$\pm$} 0.6	&66.1 {$\pm$} 1.1	&39.4 {$\pm$} 0.3		&47.3		 	\\ \hline
W$\to$[A, [D, C]]	&54.3 {$\pm$} 0.5	&63.3 {$\pm$} 1.3	&38.7 {$\pm$} 0.5		&46.9			\\ \hline
W$\to$[D, [A, C]]	&55.7 {$\pm$} 0.6	&65.8 {$\pm$} 1.0	&39.5 {$\pm$} 0.4		&\textbf{48.1} 	\\ \hline
W$\to$[C, [A, D]]	&54.2 {$\pm$} 0.6	&63.5 {$\pm$} 1.0	&39.5 {$\pm$} 0.4		&47.3 			\\ \hline \hline
Adaptation Tree 		&D$\to$A				&D$\to$W				&D$\to$C 				&Avg. 			\\ \hline \hline
D$\to$[A, W, C]		&52.6 {$\pm$} 0.5	&73.0 {$\pm$} 0.5	&39.2 {$\pm$} 0.6		&48.7 	\\ \hline
D$\to$[A, [W, C]]	&52.6 {$\pm$} 0.6	&71.8 {$\pm$} 0.7	&39.4 {$\pm$} 0.6		&48.5 			\\ \hline
D$\to$[W, [A, C]]	&54.0 {$\pm$} 0.6	&73.0 {$\pm$} 0.8	&39.9 {$\pm$} 0.6		&\textbf{49.6}  \\ \hline
D$\to$[C, [A, W]]	&53.0 {$\pm$} 0.6	&71.0 {$\pm$} 0.7	&38.9 {$\pm$} 0.6		&48.3			\\ \hline \hline
Adaptation Tree 		&C$\to$A				&C$\to$W				&C$\to$D 				&Avg. 			\\ \hline \hline
C$\to$[A, W, D]		&53.4 {$\pm$} 0.8 	&69.6 {$\pm$} 0.7	&61.2 {$\pm$} 0.9		&57.3			\\ \hline
C$\to$[A, [W, D]]	&53.2 {$\pm$} 0.7	&71.2 {$\pm$} 0.7	&63.0 {$\pm$} 1.1		&\textbf{57.8} 	\\ \hline
C$\to$[W, [A, D]]	&53.2 {$\pm$} 0.6	&69.5 {$\pm$} 0.6	&59.7 {$\pm$} 1.3		&57.1 		 	\\ \hline
C$\to$[D, [A, W]]	&52.4 {$\pm$} 0.6	&68.9 {$\pm$} 1.0	&61.0 {$\pm$} 1.1		&56.5  			\\ \hline
\end{tabular}

%% file: table_QDS.tex
\begin{tabular}{|l|c|c|c|c|}\hline 
   				& Amazon 		& DSLR 			& Webcam 		& Caltech256\\ \hline \hline
   Amazon 		& \textemdash	& 8.13			& 9.03			& \textbf{9.78} \\ \hline
   DSLR			& 8.13			& \textemdash	& \textbf{9.60} 	& 8.25 \\ \hline
   WebCam		& 9.03			& \textbf{9.60}	& \textemdash	& 8.96 \\ \hline
   Caltech256 	& \textbf{9.78}	& 8.25			& 8.96			& \textemdash	 \\ \hline
\end{tabular}

%% file: table_latent_domain.tex
\begin{tabular}{|l|l|l||c|c|c|c|}\hline
\multicolumn{3}{|c|}{Source Domain}   				&A              	   &W           			&D 	   				&C  					\\ \hline
\multicolumn{3}{|c|}{Original Target Domains}			 		&W, D, C           &A, D, C    			&A, W, C 			&A, W, D				\\ \hline \hline

Method	 	  &Hierarchy	       &Domain Discovery	    & 				   & 					& 					& 				 	\\ \hline
A-SSVM-ALL	 	  &\textemdash     &\textemdash 			&47.6 {$\pm$} 0.4  &45.4 {$\pm$} 0.4   	&46.5 {$\pm$} 0.5   &54.4 {$\pm$} 0.6 	\\ \hline
HA-SSVM 		  &2 Layers		   &Given 			&48.4 {$\pm$} 0.5  &47.3 {$\pm$} 0.5 	&\underline{48.7} {$\pm$} 0.6   &57.3 {$\pm$} 0.8    \\ \hline 
HA-SSVM 		  &3 Layers 		   &Given 			&\textbf{49.5} {$\pm$} 0.4  &\textbf{48.1} {$\pm$} 0.6 	&\textbf{49.6} {$\pm$} 0.5   &\underline{57.8} {$\pm$} 0.7    \\ \hline \hline

HA-SSVM 		  &2 Layers 		   &	LatDD-Pr 			&46.2 {$\pm$} 0.3  &45.2 {$\pm$} 0.4 	&45.9 {$\pm$} 0.4   &53.1 {$\pm$} 0.6    \\ \hline 
HA-SSVM   	  	  &3 Layers  	   &LatDD-Pr 			&46.3 {$\pm$} 0.4  &45.1 {$\pm$} 1.4 	&45.8 {$\pm$} 0.4   &53.0 {$\pm$} 0.7    \\ \hline \hline

HA-SSVM 		  &2 Layers 		   &Reshape-Pr			&\underline{49.0} {$\pm$} 0.6  &47.0 {$\pm$} 0.5 	&48.0 {$\pm$} 0.5   &\textbf{59.4} {$\pm$} 0.5    \\ \hline 
HA-SSVM  		  &3 Layers 		   &Reshape-Pr  			&\underline{49.1} {$\pm$} 0.6  &\underline{47.9} {$\pm$} 0.5 	&48.2 {$\pm$} 0.5   &\textbf{59.1} {$\pm$} 0.5    \\ \hline

\end{tabular}

%% file: hasvm.bbl
\begin{thebibliography}{}

\bibitem[A.Geiger et~al., 2011]{Geiger:2011}
A.Geiger, C.Wojek, and R.Urtasun (2011).
\newblock Joint {3D} estimation of objects and scene layout.
\newblock In {\em \NIPS}, Granada, Spain.

\bibitem[Aytar and Zisserman, 2011]{Aytar:2011}
Aytar, Y. and Zisserman, A. (2011).
\newblock Tabula rasa: Model transfer for object category detection.
\newblock In {\em \ICCV}.

\bibitem[Behley et~al., 2013]{Behley:2013}
Behley, J., Steinhage, V., and Cremers, A.~B. (2013).
\newblock Laser-based segment classification using a mixture of bag-of-words.
\newblock In {\em \IROS}.

\bibitem[Ben-David et~al., 2009]{Ben:2009}
Ben-David, S., Blitzer, J., Crammer, K., Kulesza, A., Pereira, F., and Vaughan,
  J. (2009).
\newblock A theory of learning from different domains.
\newblock {\em \ML}, 79(1):151--175.

\bibitem[Bergamo and Torresani, 2010]{Bergamo:2010}
Bergamo, A. and Torresani, L. (2010).
\newblock Exploring weakly-labeled web images to improve object classification:
  a domain adaptation approach.
\newblock In {\em \NIPS}, Vancouver, BC, Canada.

\bibitem[Dalal and Triggs, 2005]{Dalal:2005}
Dalal, N. and Triggs, B. (2005).
\newblock Histograms of oriented gradients for human detection.
\newblock In {\em \CVPR}, San Diego, CA, USA.

\bibitem[{Daum\'{e} \protect{III}}, 2007]{Daume:2007}
{Daum\'{e} \protect{III}}, H. (2007).
\newblock Frustratingly easy domain adaptation.
\newblock In {\em \ACL}, Prague, Czech Republic.

\bibitem[{Daum\'{e} \protect{III}}, 2009]{Daume:2009}
{Daum\'{e} \protect{III}}, H. (2009).
\newblock Bayesian multitask learning with latent hierarchies.
\newblock In {\em UAI}, Montreal, QC, Canada.

\bibitem[Doll\'ar et~al., 2012]{Dollar:2012}
Doll\'ar, P., Wojek, C., Schiele, B., and Perona, P. (2012).
\newblock Pedestrian detection: an evaluation of the state of the art.
\newblock {\em \TPAMI}, 34(4):743--761.

\bibitem[Duan et~al., 2009]{Duan:2009b}
Duan, L., Tsang, I.~W., Xu, D., and Chua, T.-S. (2009).
\newblock Domain adaptation from multiple sources via auxiliary classifiers.
\newblock In {\em \ICML}, Montreal, Quebec, Canada.

\bibitem[Duan et~al., 2012]{Duan:2012}
Duan, L., Xu, D., and Tsang, I.~W. (2012).
\newblock Learning with augmented features for heterogeneous domain adaptation.
\newblock In {\em \ICML}, Edinburgh, Scotland.

\bibitem[Ess et~al., 2007]{Ess:2007}
Ess, A., Leibe, B., and Gool, L.~V. (2007).
\newblock Depth and appearance for mobile scene analysis.
\newblock In {\em \ICCV}, Rio de Janeiro, Brazil.

\bibitem[Felzenszwalb et~al., 2010]{Felzenszwalb:2010}
Felzenszwalb, P., Girshick, R., McAllester, D., and Ramanan, D. (2010).
\newblock Object detection with discriminatively trained part based models.
\newblock {\em \TPAMI}, 32(9):1627--1645.

\bibitem[Finkel and Christopher, 2009]{Finkel:2009}
Finkel, J. and Christopher, D. (2009).
\newblock Hierarchical bayesian domain adaptation.
\newblock In {\em NAACL}, Colorado, USA.

\bibitem[Geiger et~al., 2012]{Geiger:2012}
Geiger, A., Lenz, P., and Urtasun, R. (2012).
\newblock Are we ready for autonomous driving? the kitti vision benchmark
  suite.
\newblock In {\em \CVPR}, Washington, DC, USA.

\bibitem[Girshick, 2012]{Girshick:2012}
Girshick, R. (2012).
\newblock {\em From Rigid Templates to Grammars: Object Detection with
  Structured Models}.
\newblock PhD thesis, The University of Chicago, Chicago, IL, USA.

\bibitem[Girshick et~al., 2012]{DPM:release5}
Girshick, R., Felzenszwalb, P., and McAllester, D. (2012).
\newblock Discriminatively trained deformable part models, release 5.
\newblock http://people.cs.uchicago.edu/~rbg/latent-release5/.

\bibitem[Gong et~al., 2013]{Gong:2013}
Gong, B., Grauman, K., and Sha, F. (2013).
\newblock Reshaping visual datasets for domain adaptation.
\newblock In {\em \NIPS}, Lake Tahoe, NV, USA.

\bibitem[Gong et~al., 2014]{gong_ijcv:2014}
Gong, B., Grauman, K., and Sha, F. (2014).
\newblock Learning kernels for unsupervised domain adaptation with applications
  to visual object recognition.
\newblock {\em \IJCV}, 109(1-2):3--27.

\bibitem[Gong et~al., 2012]{Gong:2012}
Gong, B., Shi, Y., Sha, F., and Grauman, K. (2012).
\newblock Geodesic flow kernel for unsupervised domain adaptation.
\newblock In {\em \CVPR}, Providence, RI, USA.

\bibitem[Gopalan et~al., 2011]{Gopalan:2011}
Gopalan, R., Li, R., and Chellappa, R. (2011).
\newblock Domain adaptation for object recognition: An unsupervised approach.
\newblock In {\em \ICCV}, Barcelona, Spain.

\bibitem[Griffin et~al., 2007]{Griffin:2007}
Griffin, G., Holub, A., and Perona, P. (2007).
\newblock Caltech-256 object category dataset.
\newblock Technical report, California Institute of Technology.

\bibitem[Hoffman et~al., 2012]{Hoffman:2012}
Hoffman, J., Kulis, B., Darrell, T., and Saenko, K. (2012).
\newblock Discovering latent domains for multisource domain adaptation.
\newblock In {\em \ECCV}, Florence, Italy.

\bibitem[Hoffman et~al., 2014]{hoffman_IJCV:2014}
Hoffman, J., Rodner, E., Donahue, J., Kulis, B., and Saenko, K. (2014).
\newblock Asymmetric and category invariant feature transformations for domain
  adaptation.
\newblock {\em \IJCV}, 109(1-2):28--41.

\bibitem[Hoffman et~al., 2013]{Hoffman:2013}
Hoffman, J., Rodner, E., Donahue, J., Saenko, K., and Darrell, T. (2013).
\newblock Efficient learning of domain invariant image representations.
\newblock In {\em \ICLR}, Arizona, USA.

\bibitem[Jiang, 2008]{Jiang:2008}
Jiang, J. (2008).
\newblock A literature survey on domain adaptation of statistical classifiers.
\newblock Technical report, School of Information Systems, Singapore Management
  University.

\bibitem[Kan et~al., 2014]{kan_ijcv:2014}
Kan, M., Wu, J., Shan, S., and Chen, X. (2014).
\newblock Domain adaptation for face recognition: Targetize source domain
  bridged by common subspace.
\newblock {\em \IJCV}, 109(1-2):94--109.

\bibitem[Kulis et~al., 2011]{Kulis:2011}
Kulis, B., Saenko, K., and Darrell, T. (2011).
\newblock What you saw is not what you get: Domain adaptation using asymmetric
  kernel transforms.
\newblock In {\em \CVPR}, Washington, DC, USA.

\bibitem[Mansour et~al., 2008]{Mansour:2008}
Mansour, Y., Mohri, M., and Rostamizadeh, A. (2008).
\newblock Domain adaptation with multiple sources.
\newblock In {\em \NIPS}, Vancouver, Canada.

\bibitem[Mirrashed and Rastegar, 2013]{Fatemeh:2013}
Mirrashed, F. and Rastegar, M. (2013).
\newblock Domain adaptive classification.
\newblock In {\em \ICCV}, Sydney Australia.

\bibitem[Mosek, 2013]{Mosek:2013}
Mosek (2013).
\newblock Optimization toolkit.
\newblock http://www.mosek.com.

\bibitem[Ni et~al., 2013]{Ni:2013}
Ni, J., Qiu, Q., and Chellappa, R. (2013).
\newblock Subspace interpolation via dictionary learning for unsupervised
  domain adaptation.
\newblock In {\em \CVPR}, Oregon, USA.

\bibitem[Pan and Yang, 2009]{Pan:2009}
Pan, S. and Yang, Q. (2009).
\newblock A survey on transfer learning.
\newblock {\em \KDE}, 22(10):1345--1359.

\bibitem[Saenko et~al., 2010]{Saenko:2010}
Saenko, K., Hulis, B., Fritz, M., and Darrel, T. (2010).
\newblock Adapting visual category models to new domains.
\newblock In {\em \ECCV}, Hersonissos, Heraklion, Crete, Greece.

\bibitem[V\'azquez et~al., 2014]{Vazquez:2014}
V\'azquez, D., L\'opez, A., Mar\'in, J., Ponsa, D., and Ger\'onimo, D. (2014).
\newblock Virtual and real world adaptation for pedestrian detection.
\newblock {\em \TPAMI}, 36(4):797--809.

\bibitem[V\'azquez et~al., 2012]{Vazquez:2012}
V\'azquez, D., L\'opez, A., and Ponsa, D. (2012).
\newblock Unsupervised domain adaptation of virtual and real worlds for
  pedestrian detection.
\newblock In {\em \ICPR}, Tsukuba, Japan.

\bibitem[Xu et~al., 2014a]{Xu_PAMI:2014}
Xu, J., Ramos, S., V{\'a}zquez, D., and L{\'o}pez, A. (2014a).
\newblock Domain adaptation of deformable part-based models.
\newblock {\em \TPAMI}.

\bibitem[Xu et~al., 2014b]{Xu_ITS:2014}
Xu, J., V{\'a}zquez, D., L{\'o}pez, A., Mar{\'i}n, J., and Ponsa, D. (2014b).
\newblock Learning a part-based pedestrian detector in a virtual world.
\newblock {\em \TITS}.

\bibitem[Yang et~al., 2007]{Yang:2007}
Yang, J., Yan, R., and Hauptmann, A. (2007).
\newblock Cross-domain video concept detection using adaptive {SVMs}.
\newblock In {\em ACM Multimedia}, Augsburg, Germany.

\bibitem[Zhu et~al., 2010]{Zhu:2010}
Zhu, L., Chen, Y., Yuille, A., and Freeman, W. (2010).
\newblock Latent hierarchical structural learning for object detection.
\newblock In {\em \CVPR}, San Francisco, CA, USA.

\end{thebibliography}
